\documentclass[lettersize,journal]{IEEEtran}
\usepackage{amsmath,amsfonts}
\usepackage{algorithm}
\usepackage{algorithmicx}
\usepackage{algpseudocode}
\usepackage{array}
\usepackage[caption=false,font=normalsize,labelfont=sf,textfont=sf]{subfig}
\usepackage{textcomp}
\usepackage{stfloats}
\usepackage{url}
\usepackage{verbatim}
\usepackage{graphicx}
\usepackage{cite}
\usepackage{fancyvrb}
\usepackage{fvextra}
\usepackage{comment}
\usepackage{booktabs} 
\usepackage{xcolor}
\usepackage{placeins}
\usepackage{rotating}

\newcommand{\tb}[1]{{\color{magenta}: #1}}

\newcommand{\ES}{(1\,$\stackrel{+}{,}\,$1)-EA}
\newcommand{\MLES}{($\mu\,\stackrel{+}{,}\,\lambda$)-EA}
\newcommand{\LLMES}{(1\,$\stackrel{+}{,}\,$1)-LLaMEA}
\newcommand{\LLMEA}{LLaMEA}

\newcommand{\reply}[3]{{\color{black} #3}}
\newcommand{\clari}[1]{{\color{black} #1}}
\newcommand{\revtwo}[1]{#1}

\begin{document}

\title{LLaMEA:\\ A Large Language Model Evolutionary Algorithm for Automatically Generating Metaheuristics}


\author{Niki van Stein,~\IEEEmembership{Member, IEEE},
        and Thomas B{\"a}ck,~\IEEEmembership{Fellow, IEEE}
\thanks{Manuscript received \today \emph{ (Corresponding author: Niki van Stein)}.
Niki van Stein (n.van.stein@liacs.leidenuniv.nl), Niki van Stein and Thomas B{\"a}ck are with the Leiden Institute of Advanced Computer Science (LIACS), Leiden University, The Netherlands.}
}



\maketitle

\begin{abstract}
Large Language Models (LLMs) such as GPT-4 have demonstrated their ability to understand natural language and generate complex code snippets. This paper introduces a novel Large Language Model Evolutionary Algorithm (\LLMEA)
framework, leveraging GPT models for the automated generation and refinement of algorithms.
Given a set of criteria and a task definition (the search space), \LLMEA\ iteratively generates, mutates and selects algorithms based on performance metrics and feedback from runtime evaluations. This framework offers a unique approach to generating optimized algorithms without requiring extensive prior expertise. 
We show how this framework can be used to generate novel 
black-box metaheuristic optimization algorithms 
\revtwo{for box-constrained, continuous optimization problems}
automatically. 
\LLMEA\ generates multiple algorithms that outperform state-of-the-art optimization algorithms (Covariance Matrix Adaptation Evolution Strategy and Differential Evolution) on the 5-dimensional black box optimization benchmark (BBOB). 
The algorithms also show competitive performance on the 10- and 20-dimensional instances of the test functions, although they have not seen such instances during the automated generation process. 
The results demonstrate the feasibility of the framework and identify future directions for automated generation and optimization of algorithms via LLMs. 
\end{abstract}


\IEEEpeerreviewmaketitle

\section{Introduction}\label{sec:introduction}

\IEEEPARstart{F}{or} decades, algorithms for finding near-optimal solution candidates to global optimization problems 
\clari{of the form 
\begin{equation}\label{eqn:optimization}
    \text{minimize} \; \mathcal{F}: \mathcal{S} \rightarrow \mathbb{R}
\end{equation}
where $\mathcal{S} \subseteq \mathbb{R}^d$ and $\mathcal{S} = \times_{i=1}^d [l_i,u_i]$ is defined by box constraints ($l_i < u_i$, $\forall \, l_i, u_i \in \mathbb{R}$),
}
have been developed based on inspirations gleaned from nature.
Famous examples include the use of biological evolution in the field of Evolutionary Computation \cite{eiben2015introduction,back1997handbook} (with examples such as Genetic Algorithms and Evolutionary Strategies), the swarming behavior of bird-like objects in Particle Swarm Optimization \cite{KESwarmIntelligence}, or the foraging behavior of ants in Ant Colony Optimization \cite{Dorigo2004ACO}.
For the highly advanced variants of such algorithms, impressive results have been reported for solving real-world optimization problems \cite{greiner2022evolutionary,doi:10.1142/9781848166820_0006,chiong2012variants}.

However, the number of metaphor-inspired algorithms that have been proposed by researchers, often as relatively small variations of existing methods or claimed as a completely new branch, is very large (e.g., \cite{ma2023performance}, and \cite{del2021more} mentions more than 500 methods).
A systematic empirical benchmarking of such methods against the state-of-the-art, as exemplified in  \cite{10.1145/3638529.3654122}, is typically not performed. 

Realizing that the laborious, expert driven approach for improving existing and inventing new algorithms has become quite inefficient, researchers have recently started to develop modular frameworks for arbitrarily combining components (\emph{modules}) from algorithm classes into new variants - thereby creating combinatorial algorithm design spaces in which thousands to millions of algorithms can be generated, benchmarked, and optimized. 
Examples include the modular Covariance Matrix Adaptation Evolution Strategy (CMA-ES) \cite{modES16,modCMA23}, modular Differential Evolution \cite{modDE23}, modular Particle Swarm Optimization \cite{PSOX22}, and a recent overview that summarizes all such approaches towards automated design \cite{Design23}. 

Although they often provide algorithm design spaces of millions of potential module combinations plus their hyper-parameter search spaces, modular frameworks also need to be created by experts in the field, by carefully selecting the included modules and providing the full infrastructure for configuring and searching through such algorithm spaces. 
Moreover, the modular approaches are limited to the design space created by the choices that have been made by the experts when selecting modules for inclusion. 

In this work, we propose to overcome such limitations by using Large Language Models (LLMs) within an evolutionary loop for automatically and iteratively evolving and optimizing the program code of such metaheuristic optimization algorithms \clari{for solving the optimization problem in Equation (\ref{eqn:optimization}}). 
To evaluate the generated algorithms, we use a specific optimization benchmarking tool
\clari{(IOHprofiler, consisting of the IOHexperimenter  \cite{denobel2022iohexperimenterbenchmarkingplatformiterative} module
for systematically running algorithms on benchmark functions, and IOHanalyzer \cite{10.1162/evco_a_00342} for statistically analyzing the results of the runs)}
to automatically evaluate the quality of the generated optimization algorithms and to provide a corresponding feedback to the LLM. 
The specific contributions of this work are as follows:
\begin{itemize}
    \item We present the LLM-Evolutionary Algorithm (\LLMEA), an \clari{evolutionary} algorithm that uses an LLM to automatically generate and optimize high-quality metaheuristic optimization algorithms 
    \reply{1}{1}{for solving the optimization problem as stated in Equation~(\ref{eqn:optimization})}.
    \revtwo{\LLMEA\ combines in-context learning, a precise task prompt (including a sample algorithm, code interface requirements, and mutation instruction), error handling and two different selection strategies in a novel way.}
    \revtwo{To the best of our knowledge, this is the first time that new metaheuristics for continuous optimization problems are successfully generated and optimized by LLMs to reach and exceed state-of-the-art optimization algorithm performance levels.}
        \reply{1}{1}{In particular, we show that the generated algorithms are competitive with \revtwo{three baseline algorithms, namely (i)} the CMA-ES with default parameters, \revtwo{(ii)} a hyper-parameter-optimized CMA-ES, and \revtwo{(iii)} Differential Evolution.}
    \item  We couple \LLMEA\ with the benchmarking tool, IOH\-experimenter \cite{10.1162/evco_a_00342}, to automatically evaluate the quality of the generated optimization algorithm, for providing feedback to the LLM.
    \reply{1}{1}{IOHexperimenter is a well-established benchmarking framework that supports evaluating optimization heuristics for continuous optimization problems; see e.g.~\cite{DBLP:conf/cec/NeumannNQDNVAYWB23,DBLP:journals/asc/DoerrYHWSB20,DBLP:journals/corr/abs-1810-05281,DBLP:conf/gecco/DoerrWVBNY23,DBLP:journals/telo/WangVYDB22}. 
    It allows for comparing a new optimization algorithm automatically against a well-known set of state-of-the-art algorithms on a wide set of benchmark functions in a statistically sound way. 
    It is only this automated benchmarking approach that allows us to close the evolutionary loop that includes the LLM and to run it automatically.}
    \revtwo{We specifically use a standard set of 24 benchmark test problems (with multiple instances of each problem), the so-called black-box optimization benchmark BBOB \cite{hansen2010black}, for representing the class of continuous optimization problem as stated in Equation~(\ref{eqn:optimization}).}
    \item \revtwo{To evaluate the quality of a newly generated metaheuristic across a set of benchmark functions by a single, aggregated performance measure, we use the \emph{Area Over the Convergence Curve} ($AOCC$) measure \cite{eafecdf}.
    This enables the LLM to generate metaheuristics that perform well across a set of problem instances, rather than on a single problem, only.}
\end{itemize}

\reply{4}{2}{We also would like to emphasize that the goal of this research is not to generate a specific algorithm for combinatorial or constraint optimization problems, but rather to combine the use of modern, sound benchmarking techniques for measuring the performance of algorithms with the generative capabilities of \LLMEA\ for achieving reliable results for the continuous optimization task defined in Equation (\ref{eqn:optimization}) and for outperforming state-of-the-art metaheuristics on such problems.}

In the remainder of the paper, we first discuss the state-of-the-art in LLM-based algorithm generation for direct, global optimization (Section \ref{sec:related}). 
We then introduce the newly proposed \LLMEA\ in Section \ref{sec:llm-es}, and our experimental setup in Section \ref{sec:experiments}. 
The experimental results are presented and discussed in Section \ref{sec:results}.
We also analyze the best algorithms found in Section \ref{sec:analysis}, as we are striving to understand the resulting algorithms proposed by  \LLMEA, and why they might perform so well. 
The conclusions are presented in Section \ref{sec:conclusions}.

\section{Related Work}\label{sec:related}

The integration of Large Language Models (LLMs) into the optimization domain has recently received significant attention, resulting in various innovative methodologies. \reply{4}{2}{We can distinguish three classes of solutions that integrate LLMs and Evolutionary Computation (EC); \emph{Prompt optimization} by EC methods, \emph{LLMs as EC} method and optimization of \emph{code generation}}.

\textbf{Prompt optimization}: 
Using optimization algorithms, such as Genetic Algorithms for optimizing prompts (\textsc{EvoPrompt}) \cite{guo2024connecting} has shown impressive results in outperforming human-engineered prompts.
This idea of directly evolving the prompts by an Evolutionary Algorithm has also recently been used to demonstrate an automated engineering design optimization loop for finding 3D car shapes with optimal aerodynamic performance \cite{rios2023large}.

These approaches are limited, however, as they are usually based on a fixed prompt template and a finite set of strings as building blocks and need evolutionary operators that work on such patterns and strings.
To overcome such limitations, an LLM can be used to generate the prompts for an LLM, by asking the prompting LLM to propose and iteratively improve a prompt, based on the (quantifiable) feedback concerning the quality of the answer generated by the prompted LLM.
The \emph{Automated Prompt Engineer (APE)} \cite{zhou2022large} implements this loop by employing an iterative Monte Carlo search approach for prompt generation, in which the LLM is asked to generate new prompts similar to those with high scores. 
The authors illustrate the benefits of APE on a large set of instruction induction tasks \cite{honovich-etal-2023-instruction} and by improving a "Chain-of-Thought" prompt ("Let's think step by step") from \cite{Kojima2022LargeLM}.

\textbf{LLMs as EC}: The direct application of LLMs as Evolution Strategies for optimization tasks represents another innovative approach, named \textsc{EvoLLM}  \cite{lange2024large}. 
In this proposal, the LLM generates the means of uni-variate normal distributions which are then used as sampling distributions for proposing solution vectors for black-box optimization tasks.
To guide the LLM towards improvements, the best $m$ solution vectors from the best $k$ generations, each, are provided to the LLM, and it is prompted to generate the new mean values of the sampling distribution. 
A method such as EvoLLM leverages the LLM to perform the sampling based on the information summarized above, and it shows reasonable performance on a subset of eight of the BBOB set of benchmarking functions for continuous optimization \cite{hansen2009real} for one 5-dimensional instance of each problem. 
\textsc{EvoLLM} uses the idea of \emph{in-context learning}  \cite{luo2024incontext,dong2023survey}, which works by providing 
a set of examples and their corresponding scores to the LLM, which is then prompted to generate a score for a newly presented example.

\textbf{Code generation}: 
The recently proposed \emph{FunSearch} approach (searching in the function space) \cite{FunSearch2024} closes the loop and combines the LLM with a systematic evaluation of the quality of the generated programs, resulting in new solution algorithms for combinatorial optimization problems (cap-set problem, bin packing) after 2.5 million calls of the LLM. 
In this approach, a distributed island-based Evolutionary Algorithm is used for maintaining diversity and prompt construction based on the already generated programs. 
In the \reply{2}{5}{approach called} \emph{Algorithm Evolution using Large Language Model (AEL)} \cite{liu2023algorithm} \reply{3}{6}{and also the \emph{Evolution of Heuristics (EoH)} \cite{fei2024eoh}, 
ideas from Evolutionary Algorithms are used explicitly to design novel optimization heuristics. 
AEL and EoH treat each algorithm as a solution candidate in a population, evolving them by asking the LLM to perform crossover and mutation operators on the heuristics generated by the LLM.}
These approaches have shown superior performance over traditional heuristic methods and domain-specific models in solving small instances of the Traveling Salesperson Problem (TSP).
\reply{4}{1}{The AEL approach, however, is very specific to the TSP and cannot be applied (trivially) to the continuous optimization problem as in Equation (\ref{eqn:optimization}), which is the subject of our work.}
\reply{4}{1}{The EoH approach can be generalized to the continuous optimization problem with some small modifications, and thus be compared with our proposed approach. Note that the EoH approach was designed to generate and optimize small pieces of code (single functions) and the black-box optimization algorithms we aim to generate can be complex, relatively large and consisting of a complete class with variables and functions.}

\begin{figure*}[!ht]
    \centering
    \includegraphics[width=\textwidth, trim=5mm 5mm 5mm 5mm,clip]{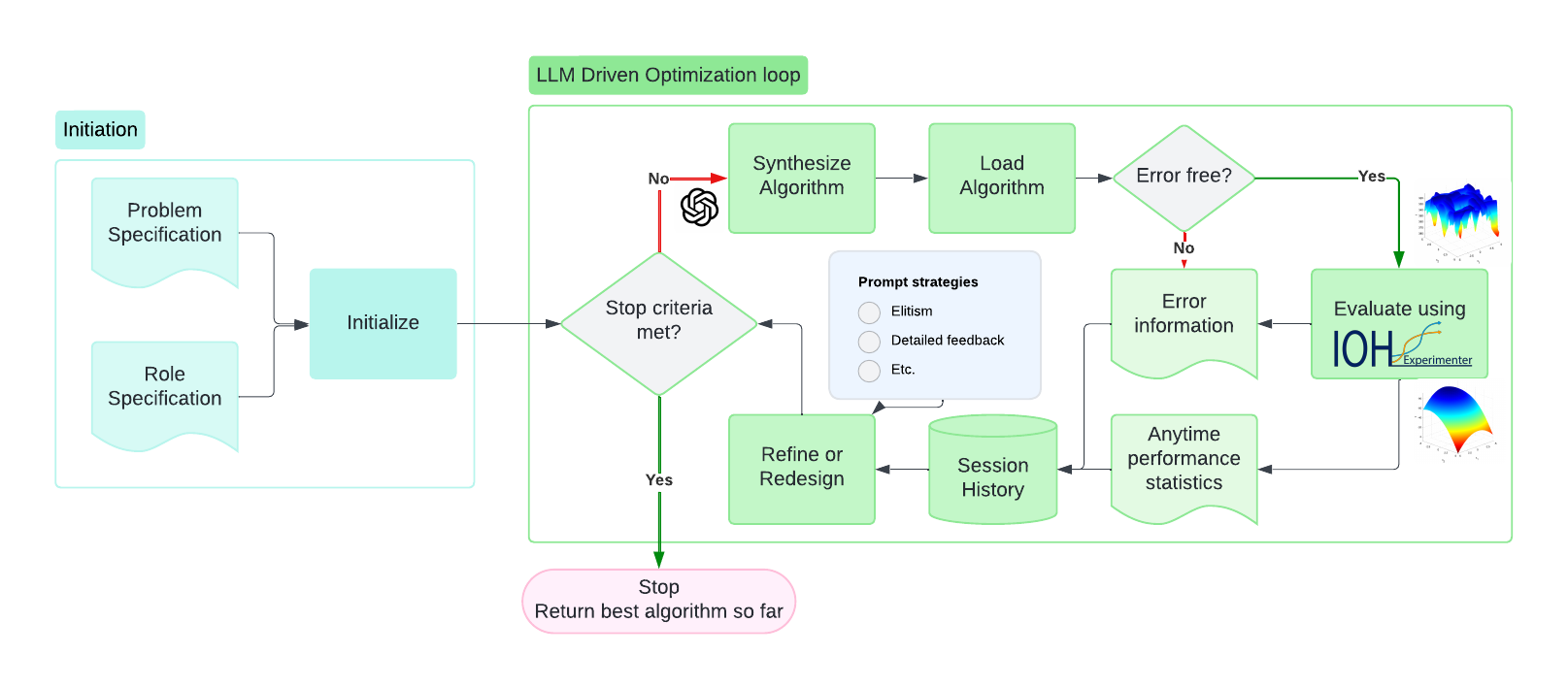}
    \caption{The summary of the proposed LLM driven algorithm design framework \LLMEA. Full details of all steps are provided in the corresponding sections. \label{fig:framework}}
\end{figure*}

The concept of recursive self-improvement, where systems iteratively improve their performance by refining their own processes, has been a focal point in AI research. 
The Self-Taught Optimizer (STOP) \cite{zelikman2024selftaught} framework exemplifies this by using LLMs to recursively enhance code generation. 
STOP begins with a seed improver that uses an LLM to generate and evaluate candidate solutions, iteratively refining its own scaffolding to improve performance across various tasks. 
This framework highlights the potential of LLMs to act as meta-optimizers, capable of self-improvement without altering their underlying model parameters.

When it comes to leveraging LLMs for the design of black-box optimization metaheuristics, results from a manual prompting approach with GPT-4 for selecting, explaining and combining components of such algorithms have illustrated the capabilities of LLMs for generating such algorithms \cite{PluhacekLLMmetaheuristics2023}. 
Automated generation and evaluation of their performance was not performed, however.

Overall, these advancements illustrate the diverse use of LLMs in an optimization context. 
However, we observe that the application of LLMs for generating novel direct global optimization algorithms from scratch is still at a very early stage. 
Often, the performance evaluation of the LLM (e.g., in \textsc{EvoLLM}) or the generated algorithms (e.g., in AEL) focuses on small test problems or does not benchmark against state-of-the-art algorithms. 
In-context learning, as in \textsc{EvoLLM}, and algorithm generation, as in STOP or AEL, are usually not combined, and the generation  of new algorithms is typically based on mutation and crossover requests to the LLM, with a focus on solving combinatorial problems.  

In our approach, described in Section \ref{sec:llm-es}, we propose to overcome such limitations by presenting a framework that can be used to generate any kind of algorithm as long there is a way to assess its quality automatically and it stays within the token limits of current LLMs.
To achieve this goal, we combine in-context learning and an Evolutionary Algorithm-like iteration loop that allows the LLM to either refine (i.e., \emph{mutate}) the best-so-far algorithm or redesign it completely. \reply{3}{9}{The proposed approach not only uses numerical feedback scores (as AEL and EoH do), but also debugging information or other textual feedback that can be automatically provided to steer the search towards better solutions.}
We combine this approach with a sound performance evaluation of the generated algorithms, based on the well-established IOHprofiler benchmark platform.


\section{\LLMEA}\label{sec:llm-es}

The proposed \clari{\emph{Large Language Model Evolutionary Algorithm}}
(\LLMEA) framework (Figure \ref{fig:framework}) consists of four primary steps, which are iteratively repeated, following a similar structure as an Evolutionary Algorithm with one parent and one child with an \emph{initialization} step and 
a general optimization loop of \emph{evaluation} (in this case, by using IOH\-experimenter \cite{IOHexperimenter}
\clari{to run the generated algorithms on a well defined set of benchmarking test functions}) and \emph{refinement (mutation) or redesign} of the algorithm. The \emph{selection strategy} determines whether only improvements are accepted (in case of a $(1+1)$-strategy) or the newly generated algorithm is always accepted (in case of a $(1,1)$-strategy).
\reply{3}{3}{Since both selection strategies are integrated into \LLMEA, we use the term \ES\ for describing both selection strategies in one notation.}

\begin{algorithm*}
\caption{\LLMES\label{alg3}}
\begin{algorithmic}[1]
\State $S \gets $ task-prompt
\Comment{Task description prompt}
\State $F_0 \gets $ task-feedback-prompt
\Comment{Feedback prompt after each iteration}
\State $t \gets 0$
\State $a_t \gets LLM(S)$
\Comment{Initialize by generating first parent program}
\State $(y_t, \sigma_t, e_t) \gets f(a_t)$
\Comment{Evaluate mean quality and std.-dev.~of first program and catch errors if occurring}
\State $a_b \gets a_t; y_b \gets y_t; \sigma_b \gets \sigma_t; e_b \gets e_t$
\Comment{Remember best-so-far}
\While{$t < T$}     
    \Comment{Budget not exhausted}
    \If{$(1+1)$}
    \Comment{$(1+1)$-Variant: Construct new prompt, using best-so-far algorithm}
    \State $F \gets (S, ((\text{name}(a_0), y_0),\ldots,(\text{name}(a_t), y_t)), (a_b, y_b, \sigma_b, e_b), F_0)$ 
    \Else 
    \Comment{$(1,1)$-Variant: Construct new prompt, using latest parent algorithm}
    \State $F \gets (S, ((\text{name}(a_0), y_0),\ldots,(\text{name}(a_t), y_t)), (a_t, y_t, \sigma_t, e_t), F_0)$ 
    \EndIf
        \State $a_{t+1} \gets LLM(F)$ 
    \Comment{Generate offspring algorithm by mutation}
    \State $(y_{t+1}, \sigma_{t+1}, e_{t+1}) \gets f(a_{t+1})$
    \Comment{Evaluate offspring algorithm, catch errors}
    \If{$e_{t+1} \neq \emptyset$} 
        $y_{t+1} = 0$ \Comment{Errors occurred}
    \EndIf
    \If{$y_{t+1} \geq y_t$} 
        $a_b \gets a_{t+1}$; $y_b \gets y_{t+1}$; $\sigma_b \gets \sigma_{t+1}$; $e_b \gets e_{t+1}$ 
        \Comment{Update best}
    \EndIf
    \State $t \gets t+1$ \Comment{Increase evaluation counter}
\EndWhile
\State \textbf{return} $a_b, y_b$ 
\Comment{Return best algorithm and its quality}
\end{algorithmic}
\end{algorithm*}


Our approach, which we therefore abbreviate as \LLMES, is presented in more detail in Algorithm \ref{alg3}. 
Here, initialization is performed in lines 1-6 by prompting the LLM with a task description prompt $S$
\clari{(see Section \ref{subsec:starting_prompt} for a description of $S$)}, evaluating the quality $f(a_t)$ of the generated program code (text) $a_t \in \mathcal{A}$, and memorizing the initial program $a_t$, its mean quality $y_t$ and standard deviation of quality $\sigma_t$ over multiple runs, and potentially some runtime error information $e_t$ as  best solution $(a_b, y_b, \sigma_b, e_b)$ 
(see Section \ref{subsec:evaluation} for a variation of this approach, in which more detailed information is provided to \LLMEA).

The while-loop (line 7-20) iterates through $T$ (total budget) LLM calls to generate a new program each time (line 13).
The prompt construction in lines 9 (in case of elitist $(1+1)$-selection) or line 11 (in case of non-elitist $(1,1)$-selection) is the essential step, generating a prompt $F$ 
\clari{(see Section \ref{subsec:mutation} for a description of the template of $F$)}
that consists of a concatenation\footnote{We use tuple notation here for clarity  to show the components that compose the prompt string.} of the following information:
(i) Task description $S$; (ii) the algorithm names and mean quality values of all previously generated algorithms; (iii) the current best algorithm's $a_b$ (in case of $(1+1)$-selection) or most recent algorithm's  $a_t$ (in case of $(1,1)$-selection) complete information (code, mean quality and standard deviation, execution error information) and (iv) a short task prompt $F_0$, with an instruction to the LLM.
The new algorithm is generated by the LLM (line 13) and its quality evaluated catching any runtime errors on the way (line 14), setting $y_{t+1}$ explicitly to zero (worst possible quality) if runtime errors occurred (line 15). 
The best-so-far algorithm is updated if an improvement was found (line 17).

It should be noted that in the general description of Algorithm \ref{alg3} given above, we do not specify the quality measure $f: \mathcal{A} \rightarrow \mathbb{R}_{\geq 0}$; $f(a) \rightarrow \max$ in detail, as \LLMEA\ is a generic concept. 
In Section \ref{subsec:evaluation}, we define the specific quality measure used here for generating metaheuristics. 

\subsection{Starting Prompt}\label{subsec:starting_prompt}

The initialisation of the optimization loop is crucial for the framework to work, as it sets the boundaries and rules that the LLM needs to operate with. Through experimentation, we found that including a small example code 
\clari{into the task prompt $S$}
helps a lot in generating code without syntax and runtime errors. However, the example code could also bias the search towards similar algorithms. In this work we therefore choose to use a simple implementation of random search as the example code to provide. 
The code of this random search algorithm is provided in our Zenodo repository \cite{anonymous_2024_11358117}.
The LLM receives an initial prompt with a specific set of criteria, domain expertise, and problem description. This starting point guides the LLM in generating an appropriate algorithm. The starting prompt provides the role description first, followed by  a detailed description of the problem to solve including a clear format of the expected response.
Our detailed task prompt $S$ is given below: 
\vspace{2.5\baselineskip}

\begin{small}
\begin{Verbatim}[breaklines=true, frame=single, label=Detailed task prompt $S$]
Your task is to design novel metaheuristic algorithms to solve black box optimization problems.
The optimization algorithm should handle a wide range of tasks, 
which is evaluated on a large test suite of noiseless functions. 
Your task is to write the optimization algorithm in Python code. 
The code should contain one function `def __call__(self, f)`, 
which should optimize the black box function `f` using 
`budget` function evaluations.
The f() can only be called as many times as the budget allows.
An example of such code is as follows:
```
<initial example code>
```
Give a novel heuristic algorithm to solve
this task. 
Give the response in the format:
# Name: <name of the algorithm>
# Code: <code>
\end{Verbatim}
\end{small}

\subsection{Algorithm Synthesis (Initialization)}
Using the prompt \clari{$S$}, the LLM generates the code for a new metaheuristic optimization algorithm, considering the constraints and guidance provided.
The LLM should provide the answer in the format given, using MarkDown formatting for the code-block. The generated algorithm and its name are extracted using regular expressions, including additional exception handling to capture small deviations from the requested format. In our experiments, 100\% of the LLM responses lead to successfully extracted code. In addition, the LLM usually generates a small explanation in addition to what we ask, which we store for offline evaluation.
In the case that we cannot extract the code (which in practice never happened but can theoretically occur), we provide the LLM feedback that the response did not follow the provided format, trying to enforce the format for the next iteration.

\subsubsection*{Exception handling}
Once the algorithm is extracted we dynamically load the generated Python code and instantiate a version of the algorithm for evaluation. The loading and instantiating of the code can lead to syntax errors, which we capture and store to be provided in the refinement prompt $F$.
When these errors occur, the evaluation run cannot commence, and is therefore skipped. Evaluation metrics are set to the lowest possible value ($0$) for the particular candidate and error messages $e_t$ are stored for additional feedback to the LLM.

\subsection{Evaluation}\label{subsec:evaluation}
The generated algorithm is evaluated on the Black-Box Optimization Benchmark (BBOB) suite 
(see \cite{hansen2009real} and the supplemental material for an overview of the functions). The suite consists of $24$ noiseless functions, with an instance generation mechanism to generate a diverse set of different optimization landscapes that share particular function characteristics. 
The suite is divided into five function groups: (i) separable functions, (ii) functions with low or moderate conditioning, (iii) high conditioning unimodal functions, (iv) multi-modal functions with strong global structure and (v) multi-modal functions with weak global structure.
The BBOB suite is considered a best-practice in the field of metaheuristic algorithm design for the evaluation of newly proposed algorithms. 
For the evaluation feedback provided to the LLM, we summarize the performance on the whole BBOB suite by taking an average any-time performance metric, namely the normalized \emph{Area Over the Convergence Curve} (AOCC) (see Equation (\ref{eq:AOCC})).
This results in one number $y_t$ representing aggregated performance of the proposed algorithm over the complete benchmark suite, including multiple instances per function, and its standard deviation $\sigma_t$ over multiple runs of the algorithm.

\reply{4}{3}{Each generated algorithm is run for a fixed budget $B$ of function evaluations, and the IOHexperimenter suite makes sure that the algorithm is terminated after using the full budget of evaluations. In practice, this mechanism ensures that generated algorithms do not exceed the budget and always terminate.}
In addition to aggregating these values over all functions, we also experiment with a more detailed feedback mechanism where the average $AOCC$ and its standard deviation are returned for each function group,
resulting in ten performance values $(y_{t,1},\ldots,y_{t,5}), (\sigma_{t,1},\ldots,\sigma_{t,5})$
\clari{(two per function group)}.
Theoretically, this approach should provide the LLM with information on what kind of functions the proposed algorithm works well and on which it works less well.

\subsection{Mutation, Selection and Feedback}
\label{subsec:mutation}
The mutation and selection step in the \LLMEA\ framework consists mostly of the construction of the feedback prompt to the LLM in order to generate a new solution. 
Depending on the selection strategy, either the current-best algorithm $a_b$ is given back ($(1+1)$-strategy) to the LLM including the score it had on the BBOB suite, or the last generated algorithm $a_t$ ($(1,1)$-strategy) is provided to the LLM in the feedback prompt $F$.

After the selection is made, a feedback prompt $F$ (lines 9, 11 of Algorithm \ref{alg3}) is constructed using the following template:
\vspace*{0.5\baselineskip}

\begin{small}
\begin{Verbatim}[breaklines=true,frame=single, label=Feedback prompt template $F$]
<Task prompt S>
<List of previously generated algorithm names with mean $AOCC$ score>
<selected algorithm to refine (full code) and mean and std $AOCC$ scores>
Either refine or redesign to improve the algorithm.
\end{Verbatim}
\end{small}

The prompt includes the initial detailed task prompt $S$, a list 
$((\text{name}(a_0), y_0),\ldots,(\text{name}(a_t), y_t))$
of previously generated algorithm names and their mean scores, the selected algorithm  
$a_b$ or $a_t$, including its score $y_b$ ($y_t$) and standard deviation $\sigma_b$ ($\sigma_t$),
to mutate and a short task prompt 
$F_0 = $ \textit{"Either refine or redesign to improve the algorithm"}
telling the LLM to perform a mutation or redesign (restart) action.
The list of previously tried algorithm names is included to make sure the LLM is not generating (almost) the same algorithm twice.

The construction of the feedback prompt includes a few choices, which in general can be seen in an evolutionary computation context as follows:

\subsubsection*{Restarts and Mutation Rate}
We  ask the LLM to either make a (small) refinement of an algorithm or redesign it completely, where the latter is analogue to a restart or very large mutation rate in an Evolutionary Algorithm optimization run, increasing its exploration behaviour. 
In the proposed framework we leave this choice to the LLM itself as our task is simply \textit{"Either refine} \textit{\bf{or}} \textit{redesign to improve the algorithm"}. 
In Section \ref{subsec:mut}, we analyze how the LLM makes this decision over time, in terms of the generated algorithm names as well as code similarity.

\subsubsection*{Plus and Comma Strategy}
In the proposed framework we support both $(1+1)$- and $(1,1)$-\LLMEA\ strategies, meaning the LLM either is asked to refine/redesign the best-so-far algorithm (in the elitist $(1+1)$-case), or the last generated algorithm (in the $(1,1)$-case). 
Only the full code of the selected (best or last) algorithm is provided to the LLM in every iteration. 
In our experiments we demonstrate the differences between both strategies per LLM model.

\subsubsection*{Detailed feedback}
Instead of providing an overall score and standard deviation, we can provide the LLM with more performance details of the generated algorithms. 
In our case, we can provide additional metrics per BBOB function group, potentially giving the LLM information on what kind of functions the solution works well and on which ones it does not work well. 
This results in mean $AOCC$ values and standard deviations for each of the five function groups, i.e., $(y_{t,1},\ldots,y_{t,5}), (\sigma_{t,1},\ldots,\sigma_{t,5})$.

In our experiments we cover the inclusion of such details versus leaving them out. 
\revtwo{However, since adding such details never showed an increase in performance, we have left these results in the supplemental material to keep our main results concise and clear.}



\subsubsection*{Session history}
In the proposed framework we do not keep the entire run history in every iteration (including all codes $(a_0,\ldots,a_t)$), as this becomes more and more expensive as the list grows. 
However, the inclusion of a larger set of previous attempts by providing the previous solutions in code with their associated scores could in theory be beneficial. 
The LLM would be able to do multi-shot in-context learning, and potentially learn from previously generated solutions. 
This would translate to evolutionary computation terms, as to keeping an archive of best solutions or the use of predictive machine learning models in machine learning assisted optimization \cite{cheng2018model,khaldi2023surrogate}.

Instead of providing all codes, we keep a condensed list of algorithm names (generated by the LLM) and their respective scores, $((\text{name}(a_0), y_0), \ldots, (\text{name}(a_t), y_t))$, to facilitate in-context learning of the LLM \cite{luo2024incontext,dong2023survey}. 
The purpose of keeping this list and providing it to the LLM is two-fold, namely
(i) the LLM could learn what kind of algorithms work well and which work less good (by analyzing the algorithm names only), and (ii) it makes it less likely that the LLM is generating the same algorithm twice.

\section{Experimental Setup}\label{sec:experiments}
To validate the proposed evolution framework for generating and optimizing metaheuristics, we designed a set of experiments to compare three different LLMs and the two different selection strategies for each of them.
\revtwo{We then compare the best of these six combinations of LLM and selection strategy with the state of the art EoH algorithm (using the best-performing LLM) and a random search baseline (also using the best-performing LLM).} 
In addition, the best generated algorithms are analyzed and compared to several state-of-the-art baselines on the BBOB suite using additional instances 
and additional dimension settings.


\subsection{Large Language Models}
In our experiments we have limited the number of LLMs to the ChatGPT family of models, \reply{1}{5}{including gpt-3.5-turbo-0125 \cite{openai2022chatgpt35turbo}, gpt-4-turbo-2024-04-09 \cite{openai2023chatgpt4turbo} and the recently released gpt-4o-2024-05-13 \cite{openai2023chatgpt4o}. The \LLMEA\ framework leverages the OpenAI chat completion API call for querying the LLM. Each model was run with default parameters (top\_p equal to 1) and a temperature of $0.8$.}
For abbreviating LLM-names, we drop the extension "turbo" in the following.

\subsection{Benchmark Problems}

As explained before, we use the BBOB benchmark function suite \cite{hansen2009real}
within IOHexperimenter.
For a robust evaluation we use $3$ different instances per function, where an instance of a BBOB function is defined by a series of random transformations that do not alter the global function characteristics. 
In addition, we perform $3$ independent runs per function instance with different random seeds (giving a total of $9$ runs per BBOB function). Each run has an evaluation budget of $B = 10\,000$ function evaluations.
In our experiments we set the dimensionality of the optimization problems to $d=5$.
We run the main \LLMES\ optimization loop in Algorithm \ref{alg3} for $T = 100$ iterations.

\subsection{Performance Metrics}
To evaluate the generated algorithms effectively over a complete set of benchmark functions we use a so-called \emph{anytime performance measure}, meaning that it quantifies performance of the optimization algorithm over the complete budget, instead of only looking at the final objective function value. For this we use the normalized \emph{Area Over the Convergence Curve} ($AOCC$), as introduced in \cite{vanstein2024explainable}. The $AOCC$ is given in Equation (\ref{eq:AOCC}). 
\begin{equation} \label{eq:AOCC}
\hspace{-1.15em}
    \textit{AOCC}(\mathbf{y}_{a,f}) = \frac{1}{B} \sum_{i=1}^{B} \left( 1-\frac{ \min(\max((y_i), \textit{lb}), \textit{ub})  
    - \textit{lb}}{\textit{ub} - \textit{lb}} \right)
\end{equation}
Here, $\mathbf{y}_{a, f}$ is the sequence 
of best-so-far log-scaled 
\revtwo{precision values (i.e., differences $\log(y_i - f^*)$ between actual function value $y_i$ and function value $f^*$ of the global minimum of the function~\cite{hansen2022anytime})}
reached during the optimization run of algorithm $a$ on test function $f$ and $y_i$ its $i$-th component, $B=10\,000$ is the budget of function evaluations per run, $\textit{lb}$ and $\textit{ub}$ are the lower and upper bound of the function value range of interest.
Here, we use the common setting of $\textit{lb}=10^{-8}$ and $\textit{ub}=10^2$. 
Following best-practice~\cite{hansen2022anytime}, the precision values are log-scaled before calculating the $AOCC$.
The $AOCC$ is equivalent to the area under the 
so-called \emph{Empirical Cumulative Distribution Function} (ECDF) curve with infinite targets between the chosen bounds~\cite{eafecdf}. 

Following common practice, we aggregate the $AOCC$ scores of all 24 BBOB benchmark functions $f_1,\ldots,f_{24}$ by taking the mean over functions and their instances, i.e., for an algorithm $a$, 
\begin{equation}
    \textit{AOCC}(a) = \frac{1}{3 \cdot 24} \sum_{i=1}^{24} \sum_{j=1}^3
\textit{AOCC}(\mathbf{y}_{a, f_{ij}}) \; ,
\end{equation}
where $f_{ij}$ is the $j$-th instance of function $i$.
The final mean $AOCC$ over $k = 5$ independent runs of algorithm $a$ over all BBOB functions is given as feedback to the LLM in the next step, and is used as best-so-far solution in case an improvement was found.
In other words, the quality measure $f(a)$ used in Algorithm \ref{alg3} is defined as 
\begin{equation}\label{eqn:f}
    f(a) = 1/k \sum_{i=1}^k \textit{AOCC(a)} \; .
\end{equation}

In addition to this mean $AOCC$ score, any runtime or compile errors that occurred during validation are also used to give feedback to the LLM. 
In case of fatal errors (no execution took place), the mean $AOCC$ is set to the lowest possible score, zero. 

\subsection{Baselines}
\revtwo{To assess the performance of the proposed framework, we compare it with two baselines, namely 
the state-of-the-art approach \emph{Evolution of Heuristics} (EoH), and random search (RS), also evaluated on the BBOB benchmark function suite using the same performance measure based on $AOCC$, given in Equation (\ref{eqn:f}).
For EoH to work on our use-case, we modified the approach slightly, making as little changes as possible. 
We keep the same prompt as we used for the proposed approach, with the exception that it should generate a "small description" instead of a name and that it should generate "a single function with a specific name" (both required by EoH to work). 
We then linked the evaluation function to IOHexperimenter in the exact same way as for the proposed approach. 
All EoH hyper-parameters are kept at their default values.
Since EoH is minimizing by default we multiply the evaluated fitness by $-1$ before returning it to EoH. 
The random search baseline is just prompting the LLM for $100$ times using the same starting prompt. 
Both baselines use the GPT-4 LLM as this was the best-performing choice.}

\section{Results and Discussion}\label{sec:results}

\begin{figure}[!t]
    \centering
    \includegraphics[width=0.9\linewidth, trim=0mm 0mm 0mm 0mm,clip]{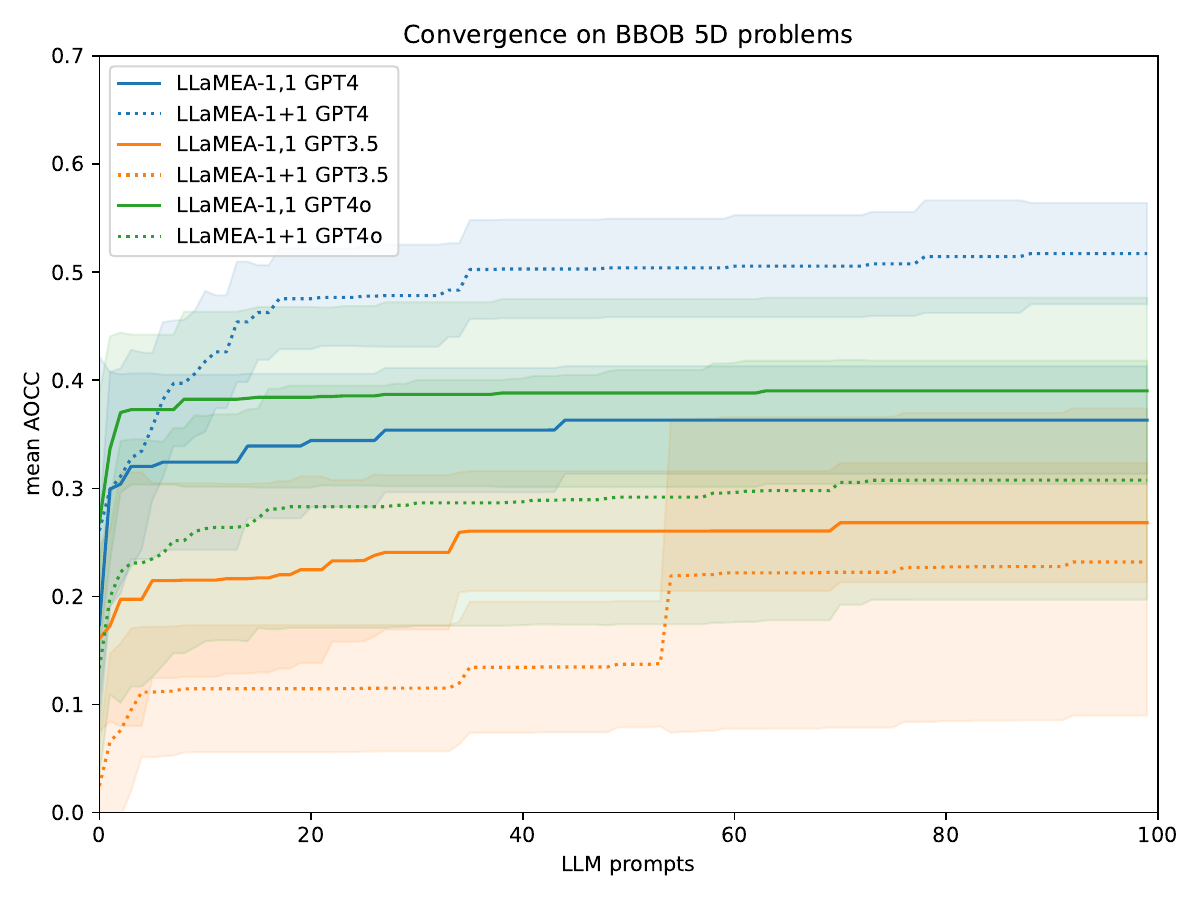}
    \caption{Mean convergence curves (best-so-far algorithm scores) over the $5$ different runs for each LLM and selection strategy. Shaded areas denote the standard deviation of the best-so-far.
    \revtwo{Please note that the difference in initial performance already results from the fact that, although the starting prompt $S$ is identical for all LLMs, the performance value shown here is the mean $AOCC$ value of the first algorithm $a_1$ generated by each LLM in line 4 of Algorithm~\ref{alg3}.
    Notice that only best-so-far values are plotted, also for (1,1)-strategies, and infeasible algorithm results are also not plotted (as they would have an $AOCC = 0$ value as mentioned in Algorithm \ref{alg3}).}
    \label{fig:convergence} }
\end{figure}

In Figure \ref{fig:convergence}, the mean best-so-far algorithm evaluation score ($AOCC$ according to Equation (\ref{eqn:f})) per configuration (LLM and strategy) is shown. 
\revtwo{Since we show only best-so-far values, all curves are monotonically increasing, also when (1,1)-selection is used.
When execution errors occur, which happens in $18.7$\% 
($843$ out of $4\,500$) 
of all runs, the corresponding $AOCC$ value of zero is not plotted, as this would make the graphs unreadable.}

\begin{figure}[!t]
    \centering
    \includegraphics[width=0.9\linewidth, trim=0mm 0mm 0mm 0mm,clip]{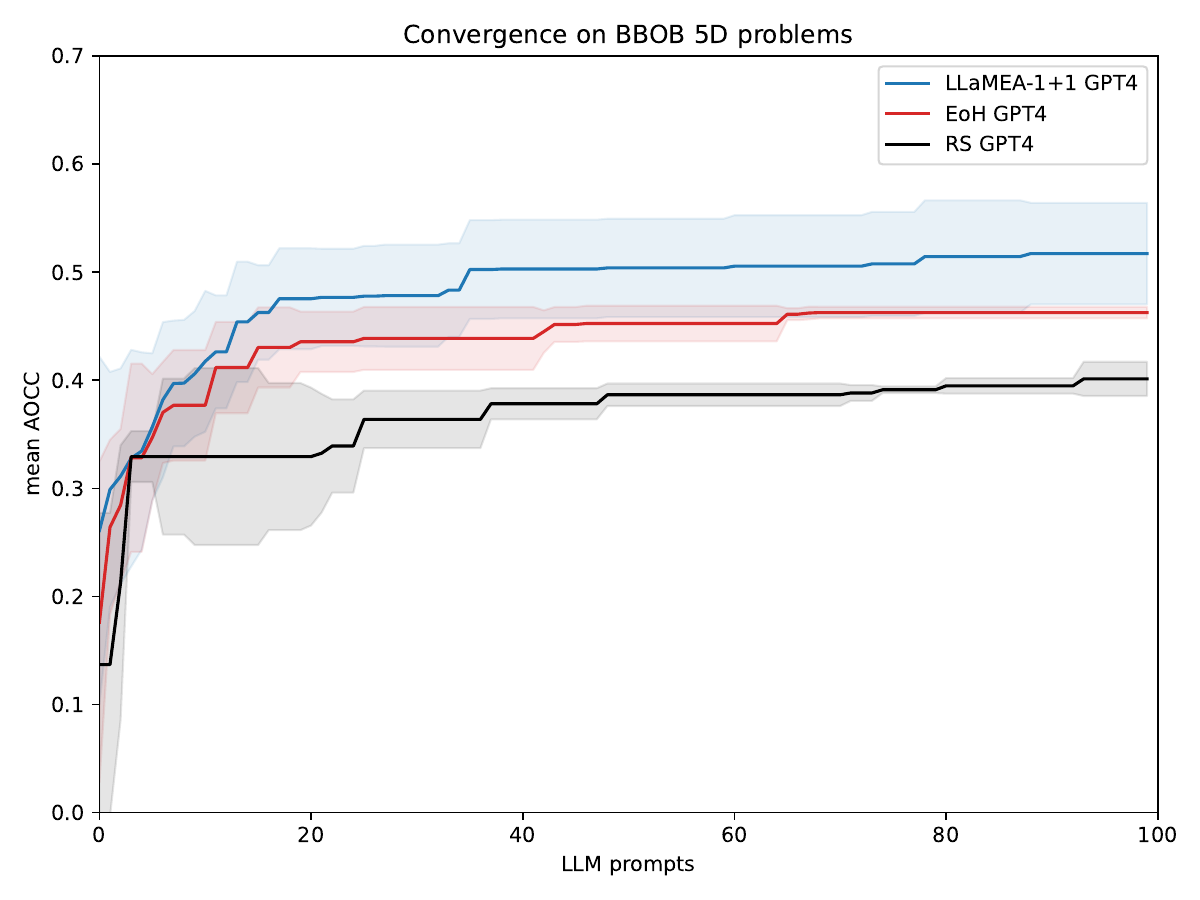}
    \caption{\revtwo{Mean convergence curves (best-so-far algorithm scores) over the $5$ different runs for the best strategy \LLMEA-variant, i.e., \LLMEA-(1+1) GPT-4 (same curve as in Fig.~\ref{fig:convergence}), including the state-of-the-art baseline EoH algorithm (red) and the random search (RS) baseline (black). 
    Shaded areas denote the standard deviation of the best-so-far over $5$ runs. 
    Additional remarks as provided in the caption of Fig.~\ref{fig:convergence} apply here, too.
    \label{fig:convergence2} }}
\end{figure}

The shaded region denotes the standard deviation over the $5$ independent runs
\clari{of Algorithm \ref{alg3} that were performed for each of the $9$ combinations of LLMs and selection operator.} 
\revtwo{Note that, of these $9$ combinations, only $6$ are shown in Figure \ref{fig:convergence}, while the other $3$ (with details) are provided in the supplemental material.}
\reply{2}{7}{This means that the commercial LLM-interfaces were called for a total of $5 \cdot 9 \cdot 100 = 4\,500$ times. Due to this costly (concerning  computational effort and funds required for using the commercial LLMs) experimentation, we have limited the number of repetitions of the runs to $5$, providing a good enough measure of the average 
performance and its variation.}
\reply{2}{9}{Since we evaluate the generated algorithms across a whole set of 24 continuous optimization problems in the BBOB test function set, and on the first three instances with three runs each, we perform $24 \cdot 9 = 216$ runs at $B=10\,000$ function evaluations each, for evaluating a single metaheuristic generated by the LLM.
This results in $2.16$ million function evaluations, and
the $4\,500$ metaheuristics generated required a total of $9.72 \cdot 10^{9}$ function evaluations.}

From Figure \ref{fig:convergence} we conclude that
different strategies yield a range of different results, depending on the model used. 
For example, the $(1+1)$-selection strategy seems to be beneficial for 
GPT-4,
but is having a deteriorating effect when using GPT-4o.
Overall 
GPT-4
seems to be better suited for the task overall, and 
GPT-3.5
is clearly a less favorable option.
\revtwo{In Figure \ref{fig:convergence2}, we compare our approach with the two baseline approaches, namely the EoH algorithm and random search (each using GPT-4, too).}
To use EoH, we extended it to work in tandem with IOHexperimenter in the same way as \LLMEA. 
\reply{4}{1}{\reply{3}{6}{The \LLMEA-$(1+1)$ GPT-4 shows better convergence (area under the $AOCC$ curve) 
than the EoH approach and also results in an overall better algorithm in the end. \revtwo{Both EoH and the proposed approach show a clear improvement over just randomly sampling the LLM, indicating the advantages of using an evolutionary search framework.}
The EoH approach follows as the second best approach, where EoH is also run $5$ times with a population size of $5$ and default hyper-parameters. 
EoH performs a bit more stable (in terms of variation over different runs), due to the larger population size. However, it suffers from tries where the generated code is not executable since it has no self-debugging capabilities (unlike \LLMEA). Another limitation of EoH is that it can only deal with single functions, while in our proposed approach complete Python classes are generated, allowing for more complex interactions.}}


\subsection{Novelty and diversity}

Analysing the generated codes and their generates names, it is immediately obvious (and not surprising) that the LLM uses existing algorithms, algorithm components and search strategies in generating the proposed solutions. 

\begin{figure}[!ht]
    \centering
    \includegraphics[width=0.9\linewidth, trim=0mm 0mm 0mm 0mm,clip]{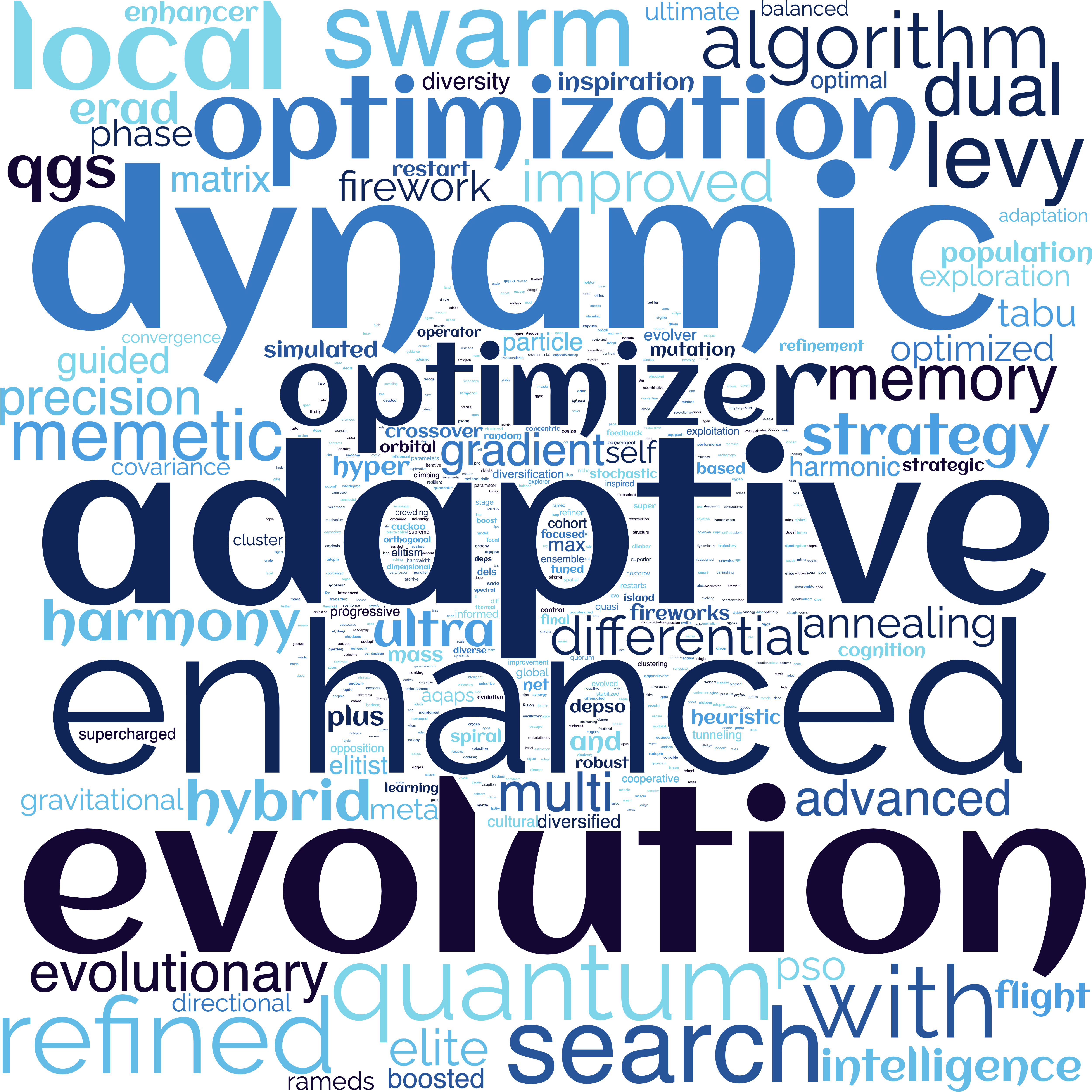}
    \caption{Word cloud of algorithm name parts generated over all different \LLMEA\ runs. }\label{fig:wordcloud} 
\end{figure}

In Figure \ref{fig:wordcloud}, a word-cloud is shown with all the sub-strings of generated algorithm names and their occurrence frequency visualized. \reply{1}{4}{Note that the word-cloud serves purely as an intuitive visual representation to quickly get an overview of the different words the algorithm names contain.
All algorithm names and codes are available in our open-source repository~\cite{anonymous_2024_11358117}.}
The generated algorithm names are in Python Camel-case style, such that it is easy to split the algorithm names into individual parts. 
In some cases, the algorithm name is just an abbreviation, and these abbreviations are kept as words in this analysis. 
The most used parts in algorithm names are rather generic, such as ``evolution", ``adaptive" and ``dynamic".
Some of the parts directly refer to existing algorithms, such as ``harmony" and ``firework", and other strings refer to existing strategies, such as ``gradient", ``local", ``elite" etc.
In general when observing the different generated solutions, we see interesting and novel combinations of existing techniques, such as \emph{``surrogate assisted differential evolution combined with covariance matrix adaptation evolution strategies''} and \emph{``dynamic firework optimizer with enhanced local search''}. 
A full list of generated algorithms and their names is provided in our Zenodo repository \cite{anonymous_2024_11358117}.
\reply{3}{2}{The algorithm names are generated automatically by the LLM that is used by LLaMEA. 
We have performed manual checks and can confirm that the algorithm names are almost always in line with the actual code. 
This means that the LLM indeed creates a descriptive name for the generated algorithm, which is based on the key algorithmic components used in the algorithm.}

\subsection{Mutation Rates}\label{subsec:mut}

\begin{figure}[!ht]
    \centering
    \includegraphics[width=\linewidth, trim=2mm 4mm 0mm 0mm,clip]{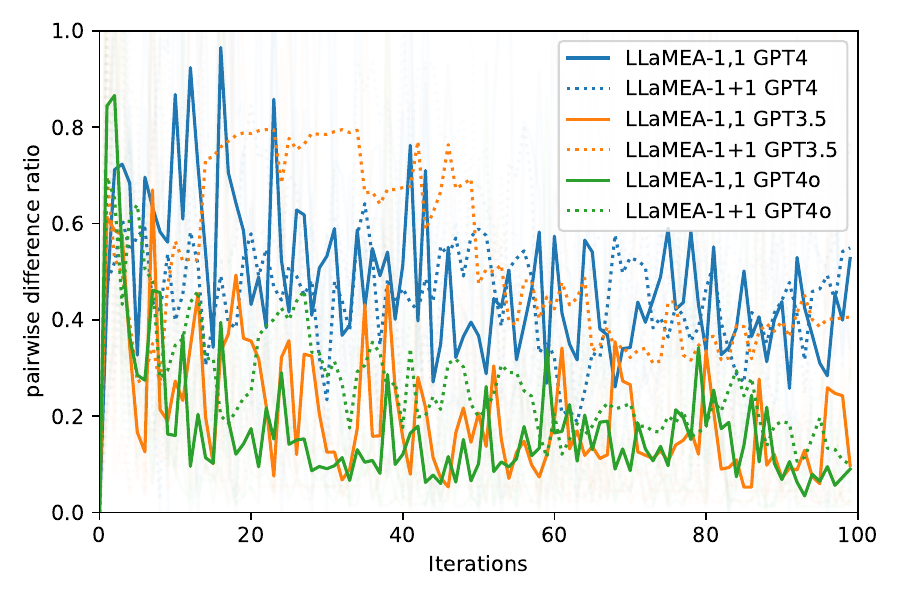}
    \caption{Pairwise differences between parent and offspring for each iteration. Solid lines represent the mean over all runs per model and strategy, more transparent lines are individual runs.\textit{w/ Details} detnotes that we use a feedback mechanism that provided not just the plain average $AOCC$ but also the average $AOCC$ per BBOB function group as feedback to the LLM.}\label{fig:diff} 
\end{figure}

To analyse how much the LLMs change (mutate) the algorithms over an entire optimization run, the code \textit{diff} ratio (number of code lines that are different between a pair of programs divided by the length of the largest code) is calculated over each run between parent and offspring solutions. In Figure \ref{fig:diff}, the mean \textit{diff} over $5$ different runs per LLM is shown. 

There are a few interesting observations we can make: 
The difference between parent and offspring for 
GPT-3.5 is on average smaller than for other models, and the 
$(1+1)$-GPT3.5-\LLMEA\
shows much higher differences in the beginning of the run than the other strategy with the same model. 
GPT-4-\LLMEA\ in general shows the largest differences (exploration) over the runs. It is very interesting to observe that the ratio of code differences in most cases seems to converge, indicating more exploration in the beginning of the search and more exploitation during the final parts of the search. 
This is interesting as the LLM has no information 
on the search budget $T$ of Algorithm \ref{alg3}, and can therefore also not base its decision to make large or small refinements on the stage of the optimization run. \reply{1}{2}{When we look at specific code-diffs between parents and offspring generated by the $(1+1)$-GPT-4-\LLMEA, we observe that the LLM mutates both hyper-parameter values and higher level logic, like introducing a new crossover or mutation operator. In addition it mutates the comments in the code to reflect and argue about the changes. See the supplemental material for some specific examples.}

We can furthermore see in the generated names of the algorithms that the LLM has either tried to refine the algorithm (adding ``Improved" or ``Refined" or ``Enhanced" to the name, or 
\clari{generating}
names such as \textless \texttt{algorithm}\textgreater V1, etc.) or redesign the algorithm using a different strategy (based on different existing algorithm names such as Differential Evolution, Particle Swarm Optimization etc.). We can therefore also look at the similarity between parent and offspring algorithm names to visualize the refinement process of the optimization runs. 

To do so, we use the \emph{Jaro similarity} \cite{jaro89}
as it gives a ratio between $0$ (completely different names) and $1$ (completely matching).
The Jaro similarity score between two algorithm names \( s \) and \( t \) is calculated as:
\begin{equation}
\text{Jaro}(s, t) = \left\{
\begin{array}{ll}
0 & \text{if } m = 0 \\
\frac{1}{3} \left( \frac{m}{|s|} + \frac{m}{|t|} + \frac{m-t}{m} \right) & \text{otherwise}
\end{array}
\right.
\end{equation}
where
\begin{itemize}
    \item \( s \) and \( t \) are the input strings.
    \item \( m \) is the number of matching characters. Two characters from \( s \) and \( t \) are considered matching if they are the same and not farther apart than \(\text{max}(|s|, |t|)/2 - 1\).
    \item \( t \) is half the number of transpositions. A transposition occurs when two matching characters are in a different order in \( s \) and \( t \).
    \item \( |s| \) and \( |t| \) are the lengths of strings \( s \) and \( t \), respectively.
\end{itemize}

\begin{figure}[!ht]
    \centering
    \includegraphics[width=\linewidth, trim=2mm 4mm 0mm 0mm,clip]{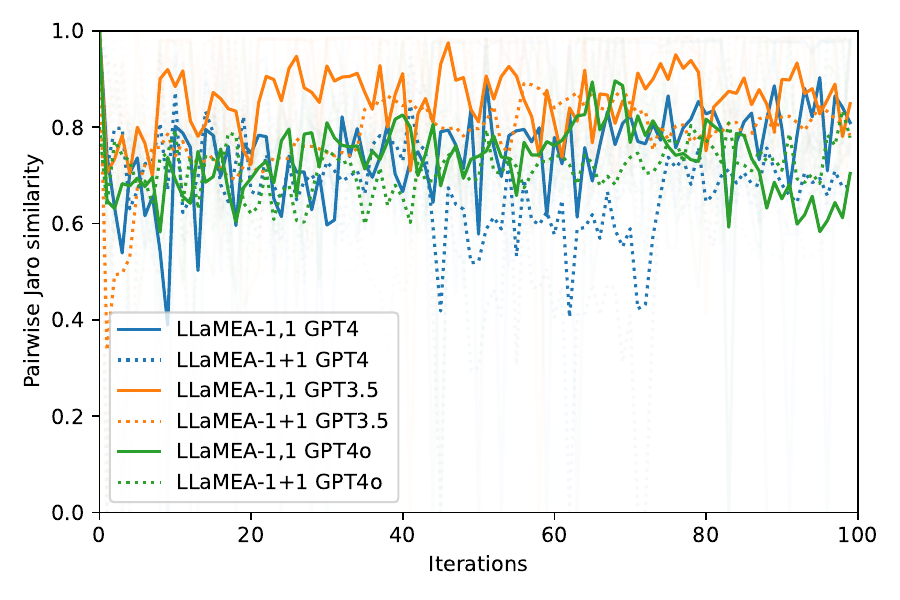}
    \caption{Average Jaro similarity scores between parent and offspring over each optimization run for different models.}\label{fig:jaro} 
\end{figure}

Figure \ref{fig:jaro} illustrates the Jaro similarity mean (five runs) of the subsequently generated algorithm names (i.e., $\text{Jaro}(\text{name}(a'), \text{name}(a_{t+1}))$ for $t = 0,\ldots,T-1$,
where $a' = a_b$ for $(1+1)$-selection and $a' = a_t$ for $(1,1)$-selection)
over iterations $t$ for each of the LLM-based optimization run configurations.

The Jaro similarity scores consistently show a behavior similar to the code difference ratios. 
It is clear that some steps only involve very small mutations (especially for GPT3.5) and sometimes a kind of restart occurs where the similarity score reaches zero in a single run.

\section{Analysis of Best Algorithms}\label{sec:analysis}

The experiment above resulted in $3\,657$ algorithms (out of a maximum\footnote{$5$ runs with $T=100$ iterations, each, for $9$ different models and strategies.} of $4\,500$) that were at least able to get an $AOCC$ score larger than zero on the BBOB suite of optimization problems in $d=5$ dimensions.
In the next step to validate our proposed framework, we evaluate how much the best of these algorithms can generalize beyond the evaluation performed during the optimization loop, and subsequently we analyze in-depth the code and behaviour of the best-performing algorithm that beats the state-the-art baselines.
The evaluation is done in $5$, $10$ and $20$ dimensions, using $5$ different instances and $5$ independent runs per instance (so $25$ runs per BBOB function to optimize).
For a fair comparison, we decided to re-run the generated algorithms for $d=5$ in this setting (with $25$ runs), too.
The state-of-the-art baselines are the Covariance Matrix Adaptation Evolution Strategy (CMA-ES)~\cite{brockhoff2012comparing}, Differential Evolution (DE)~\cite{povsik2012benchmarking} and an extensively optimized version of Modular CMA-ES \cite{modcma,vanstein2024explainable} denoted as \emph{CMA-best}. 
CMA-best is a CMA-ES algorithm with active update, a Gaussian base sampler, $\lambda=20, \mu=10$, IPOP and mirrored sampling strategy and CSA step-size-adaptation (see \cite{modcma} for the details of these configurations). 
The optimized CMA-ES algorithm is the best-configured algorithm (in terms of $AOCC$) on BBOB in $5$ dimensions and for a budget of $10\,000$ function evaluations, out of more than $52\,128$ CMA-ES and DE algorithm variants.
Beating this optimized baseline is incredibly hard as it was the result of a very large modular optimization experiment. 
We use the CMA-best baseline to obtain a reasonable upper-bound of what is possible when optimizing an algorithm configuration to a specific benchmark set and fixed budget and dimensionality.

\begin{figure*}[!ht]
    \centering
    \includegraphics[width=0.9\textwidth, trim=0mm 45mm 0mm 0mm,clip]{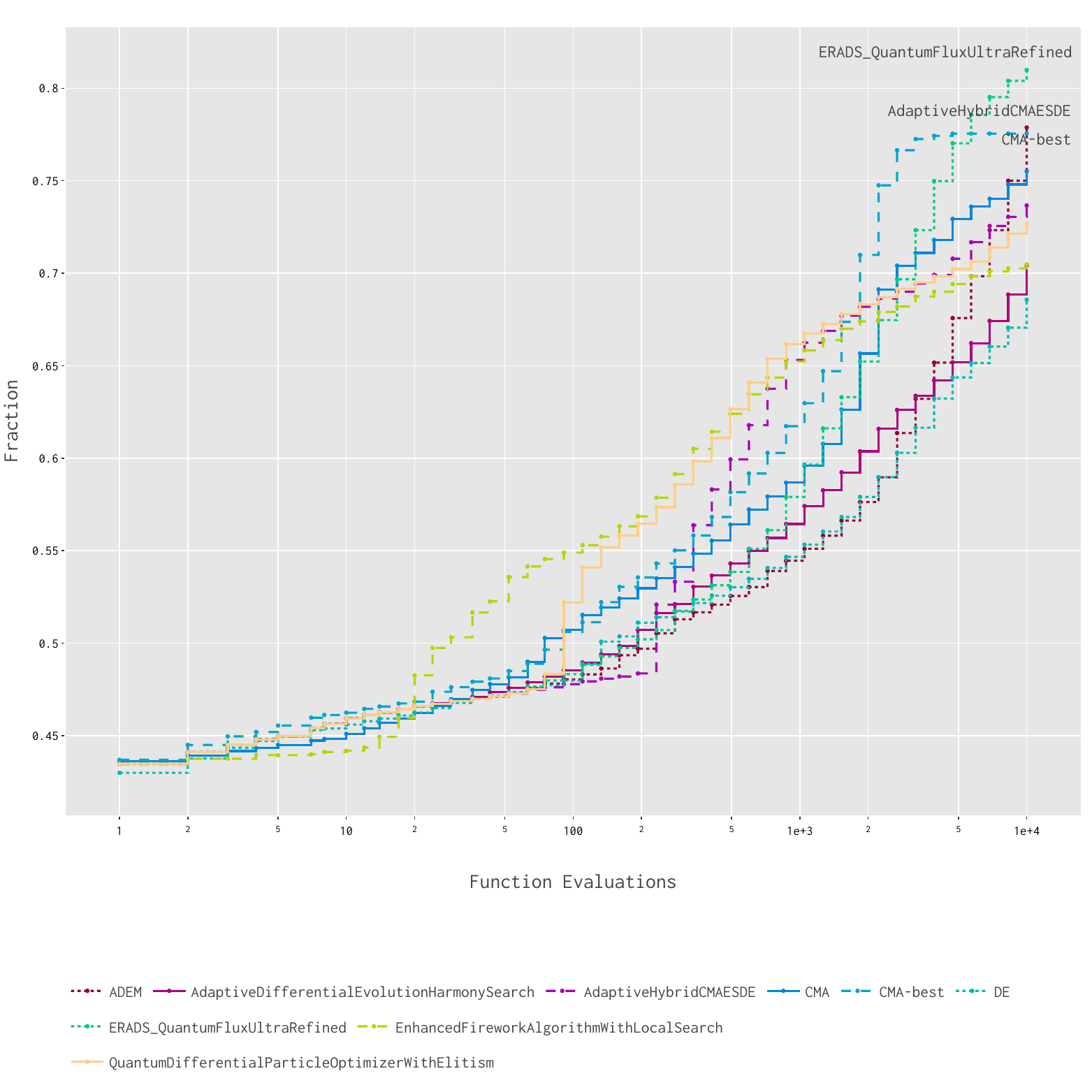}
    \includegraphics[width=0.8\textwidth, trim=20mm 0mm 0mm 235mm,clip]{plots/EAF_5d_best6.pdf}
    \caption{The empirical attainment functions (EAF) estimate the fraction of runs that attain an arbitrary target value not later than a given runtime. 
    \reply{3}{4}{The fraction ($y$-axis) denotes the cumulative fraction of target values that have been reached by an optimization algorithm, as a function of the number of function evaluations ($x$-axis).}
    We show the EAF for the best algorithms per LLM configuration (i.e., six algorithms) and the three baseline algorithms (CMA-ES, best CMA-ES, Differential Evolution (DE)), averaged over all 24 BBOB functions in $5d$.}\label{fig:ECDF} 
\end{figure*}

In Figure \ref{fig:ECDF} the \emph{empirical attainment functions} (EAF)~\cite{eafecdf}
are shown for the best algorithm generated per LLM and strategy, \reply{2}{8}{resulting in} $6$ automatically generated algorithms plus the $3$ baselines
(CMA-ES, CMA-ES-best, DE). 
The empirical attainment function estimates the fraction of runs that attain an arbitrary target value not later than a given runtime
(i.e., number of function evaluations). 
Taking the partial integral of the EAF results in a more accurate version of the Empirical Cumulative Distribution Function, since it does not rely on discretization of the targets. 
\reply{3}{4}{The "fraction" ($y$-axis in Figure \ref{fig:ECDF}) denotes the cumulative fraction of target values that have been reached by an optimization algorithm, as a function of the number of function evaluations ($x$-axis in Figure \ref{fig:ECDF}).}

Such EAF curves summarize how well an optimization algorithm performs over the complete run for the given set of target values, again on average over all instances, independent runs and objective functions. 
From the EAF curves and the area under these curves (see Table \ref{tab:AUCs}), we can observe that \LLMEA\ has found one algorithm (\emph{ERADS\_QuantumFluxUltraRefined}, for short ERADS) with better performance for $d=5$ than the state-of-the-art CMA-ES in terms of $AOCC$, two algorithms (ERADS and \emph{AdaptiveDifferentialEvolutionHarmonySearch}) that on average perform better than the optimized CMA-best after the total budget in $5d$ (the EAF curve reaches higher), two other algorithms that perform better than CMA-best in the first $1000$ evaluations (\emph{EnhancedFireworkAlgorithmWithLocalSearch} and \emph{QuantumDifferentialParticleOptizerWithElitism}) and seven algorithms that perform better than the DE baseline but worse than the CMA baseline.
Detailed convergence curves per BBOB function (for $d=5$) for ERADS\_QuantumFluxUltraRefined and three baselines are shown in Figure \ref{fig:FCE}, these curves are also available for all best algorithms in the supplemental material. From this Figure we can observe that especially for BBOB $f_{17}$ and $f_{18}$, the ERADS algorithm shows very promising search behaviour.

\begin{figure*}[!ht]
    \centering
    \includegraphics[width=0.95\textwidth, trim=8mm 40mm 0mm 0mm,clip]{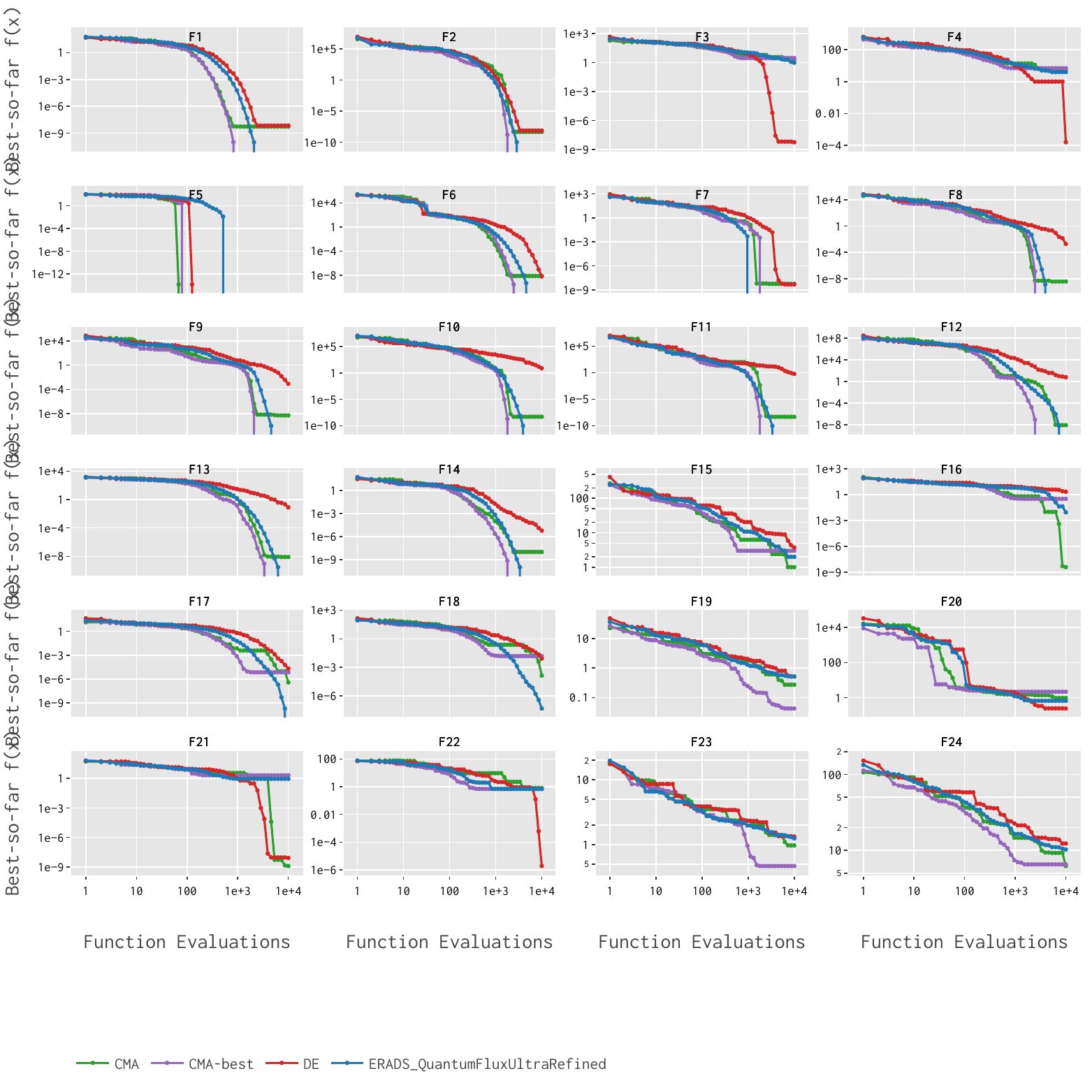}
    \includegraphics[width=\textwidth, trim=8mm 0mm 0mm 255mm,clip]{plots/FCE_5d_best1.pdf}
    \caption{Median best-so-far function value ($y$-axis) over the $10\,000$ function evaluations ($x$-axis) per BBOB function in $5d$ for the ERADS\_QuantumFluxUltraRefined algorithm and the baselines: CMA-ES, DE and CMA-best, \reply{3}{7}{for each of the 24 BBOB test functions}.}\label{fig:FCE} 
\end{figure*}

\begin{table}[!ht]
    \centering
    \caption{Area under the EAF curve scores (AUC) for the best 3 algorithms per model in $5d$, higher AUC scores stands for a better anytime performance and have a similar meaning as the $AOCC$ metric used earlier.}
    \begin{tabular}{ll}
        \textbf{ID} & \textbf{AUC} \\ \toprule
        CMA-best (optimized baseline) & 0.742 \\ 
        ERADS\_QuantumFluxUltraRefined & 0.733 \\ 
        \midrule
        CMA (baseline) & 0.703 \\ 
        \midrule
        AdaptiveHybridCMAESDE & 0.697 \\ 
        QuantumDifferentialParticleOptimizerWithElitism & 0.695 \\ 
        EnhancedFireworkAlgorithmWithLocalSearch & 0.684 \\ 
        AdaptiveHybridDEPSOWithDynamicRestart & 0.684 \\
        ADEM & 0.667 \\ 
        AdaptiveDifferentialEvolutionHarmonySearch & 0.643 \\ 
        EnhancedDynamicPrecisionBalancedEvolution & 0.641 \\
        \midrule
        DE (baseline) & 0.628 \\ 
        QPSO & 0.604 \\ 
        \bottomrule
    \end{tabular}
    \label{tab:AUCs}
\end{table}

\subsection{The "ERADS\_QuantumFluxUltraRefined" Algorithm}

\begin{algorithm}[!ht]
\caption{ERADS\_QuantumFluxUltraRefined}
\label{alg:ERADS}
\begin{algorithmic}[1]
\State $N \gets 50$ \Comment{Population size}
\State $F_{init} \gets 0.55, F_{final} \gets 0.85$ \Comment{Initial and final mutation scaling factors}
\State $CR \gets 0.95$ \Comment{Crossover probability}
\State $Memory\_factor \gets 0.3$ \Comment{Factor to integrate memory in mutation}
\State $P \gets$ u.a.r.~population initialization within $(-5.0, 5.0)$
\State $\text{fit}_P \gets f(P)$
\State $best\_index \gets \text{argmin}(\text{fit}_P)$
\State $f_{opt} \gets \text{fit}_P[best\_index]$
\State $\mathbf{x}_{opt} \gets P[best\_index]$
\State $\mathbf{\textit{Memory}} \gets \mathbf{0}$ \Comment{Initialize memory for mutation direction}
\State $t \gets N$
\While{$t < B$}
    \State $F_{current} \gets F_{init} + (F_{final} - F_{init}) \cdot \left(\frac{t}{B}\right)$
    \For{each $i$ in $[0, N-1]$}
        \State $indices \gets$ select 3  distinct indices u.a.r.~from $[0, N-1] \setminus \{i\}$
        \State $\mathbf{x1}, \mathbf{x2}, \mathbf{x3} \gets P[indices]$
        \State $\mathbf{\textit{best}} \gets P[best\_index]$
        \State $\mathbf{\textit{mutant}} \gets \mathbf{x1} + F_{current} \cdot (\mathbf{\textit{best}} - \mathbf{x1} + \mathbf{x2} - \mathbf{x3} + Memory\_factor \cdot \mathbf{\textit{Memory}})$
        \State $\mathbf{\textit{mutant}} \gets$ clip $\mathbf{\textit{mutant}}$ to bounds $(-5.0, 5.0)$
        \State $\mathbf{\textit{trial}} \gets$ crossover ($\mathbf{\textit{mutant}}, P[i]$) w.~prob.~$CR$
        \State $f\_trial \gets f(\mathbf{\textit{trial}})$
        \State $t \gets t + 1$
        \If{$f\_trial < \text{fit}_P[i]$}
            \State $P[i] \gets \mathbf{\textit{trial}}$
            \State $\text{fit}_P[i] \gets f\_trial$
            \If{$f\_trial < f_{opt}$}
                \State $f_{opt} \gets f\_trial$
                \State $\mathbf{x}_{opt} \gets \mathbf{\textit{trial}}$
                \State $best\_index \gets i$
            \EndIf
            \State $\mathbf{\textit{Memory}} \gets (1 - Memory\_factor) \cdot \mathbf{\textit{Memory}} + Memory\_factor \cdot F_{current} \cdot (\mathbf{\textit{mutant}} - P[i])$
        \EndIf
        \If{$t \geq B$}
            \State \textbf{break}
        \EndIf
    \EndFor
\EndWhile
\State \textbf{return} $\mathbf{x}_{opt}, f_{opt}$ \Comment{Return best solution and quality}
\end{algorithmic}
\end{algorithm}

While the name sounds rather futuristic and very sophisticated, upon a close inspection of the code 
the \emph{"ERADS\_QuantumFluxUltraRefined"}
algorithm looks very similar a standard Differential Evolution algorithm. 
According to the LLM, ERADS stands for \emph{"Enhanced RADEDM with Strategic Mutation"}, and RADEDM stands for \emph{"Refined Adaptive Differential Evolution with 
 Dynamic Memory"}. 
The pseudocode of \emph{"ERADS\_QuantumFluxUltraRefined"}, representing the key ideas of the generated Python code, was extracted manually and is shown in Algorithm \ref{alg:ERADS}. 
The full Python code can be retrieved from our online repository \cite{anonymous_2024_11358117}.

The main differences between ERADS and DE are the use of a memory factor to guide the mutation in certain directions and an adaptive $F$ and $CR$ control strategy.
It is unclear, why the LLM generated the "Quantum Flux" component in the algorithm's name, but it can be assumed that LLM hallucinations \cite{huang2023survey}
will also occur when applied to code generation tasks.

The proposed ERADS algorithm seems most similar to the existing JADE \cite{jade5208221} algorithm from literature. There are, however, some key differences which are, to the best of our knowledge, novel and have not been published before. 

The key differences primarily revolve around the use of a memory vector and a memory factor 
\revtwo{(lines 4, 10, 18 and 31 of Algorithm \ref{alg:ERADS})}
in combination with parameter adaptation
\revtwo{(line 13 of Algorithm \ref{alg:ERADS})}. 
\revtwo{Moreover, ERADS involves three randomly selected individuals from the population, the best individual, and the memory vector in the mutation operator (line 18) and then, with probability $CR=0.95$, performs a crossover of the mutant and the actual individual (line 20).
This crossover operator is the classical binomial crossover from DE, which copies a variable $\mathbf{v}_{i,j}$ from the mutant vector $\mathbf{v}_i$
with probability $CR=0.95$, i.e., the \emph{trial} vector in line 18 is almost always identical to the \emph{mutant} vector.
In classical DE notation, the mutant $\mathbf{v}_i$ ($i$ being the index of individuals in the population) is generated in Algorithm 2 as follows:
$$
\mathbf{v}_i = \mathbf{x}_{r_1} 
    + F_g \cdot (\mathbf{x}_{best} - \mathbf{x}_{r_1})
    + F_g \cdot (\mathbf{x}_{r_2} - \mathbf{x}_{r_3})
    + F_g \cdot m_f \cdot \mathbf{m}
$$

As usual, the $r_i$ indicate randomly sampled solution indices from the current population and $F_g$ is the mutation factor
($F_{current}$ in Algorithm~\ref{alg:ERADS}), which in ERADS changes per generation $g$. 
The first two terms and crossover can best be characterized as a DE/rand-to-best/1/bin strategy in classical DE notation \cite{XIA202133}, but the additional memory vector that contributes $0.3 \cdot F_g \cdot \mathbf{m}$ (remember that $m_f = 0.3$, as initialized in line 4, remains constant) is new.
It should also be noted that ERADS updates $\mathbf{x}_{best}$ immediately within the current generation, if the newly generated individual improves $\mathbf{x}_{best}$. 
Similarly, $\mathbf{m}$ is updated immediately, if the newly generated individual improves $\mathbf{x}_{best}$, as follows:
$$
\mathbf{m} \leftarrow 0.7 \cdot \mathbf{m} + 0.3 \cdot F_g \cdot (\mathbf{v}_i - \mathbf{x}_i)
$$

It is interesting to observe that this update does not use the \emph{trial} vector that was generated by crossover of $\mathbf{v}_i$ and $\mathbf{x}_i$ in line 20, and which is used to update $\mathbf{x}_{best}$ upon an improvement.
Instead, it just uses the mutant vector $\mathbf{v}_i$, which, however, is almost identical to the \emph{trial} vector due to} \revtwo{$CR=0.95$, which is the probability for each component of the mutant vector to be copied to the trial vector.

Finally, ERADS also includes a generational adaptation that linearly increases $F_g$ from an initial value of $0.55$ to the final value of $0.85$, i.e., $F_g = 0.55 + 0.3 \cdot t/B$ (line 13).
While this increases the mutation factor over time, it still remains below the maximum range of the mutation factor that is typically recommended, i.e., $[0,1]$ \cite{jade5208221}.
}

\revtwo{To further examine the hyper-parameters of the ERADS algorithm we include additional hyper-parameter optimization experiments in the supplemental material. The result clarifies that the LLM has already generated well performing hyper-parameter settings for $d=5$ that can not be improved further.}

In contrast to ERADS, JADE uses an optional archive with the best solutions which primarily serves to maintain diversity. Furthermore, JADE also features adaptive control parameters ($F$ and $CR$) that adjust dynamically based on the success rates of previous generations, aiming to optimise these parameters in real time to improve performance. 

It is interesting to observe that most of the best algorithms proposed by the LLM are similar to Differential Evolution (or hybrids of CMA and DE), which could indicate a bias towards this kind of algorithms in the LLM, or it could indicate that overall this kind of algorithms work well for the specific benchmark suite.

\subsection{Performance Analysis in Higher Dimensions}

The best-found algorithms using the \LLMEA\ framework are evaluated against the most relevant state-of-the-art optimizers, 
namely the previously used baselines CMA-best, CMA-ES and Differential Evolution (DE). 
CMA-ES and DE are
using recommended hyper-parameter settings.
The baselines all originate from the IOHanalyzer benchmark data set \cite{IOHanalyzer}.
In Figure \ref{fig:EAFv2}, the empirical attainment function of the proposed algorithms and baselines are shown for dimension $d \in \{10, 20\}$ using results from all BBOB functions, $5$ instances per function and $5$ random seeds ($25$ runs per BBOB function in total). 
It is interesting to observe that while ERADS performs well in $5$ dimensional problems, it fails to generalize well to higher dimensions, while \emph{EnhancedFireworkAlgorithmWithLocalSearch} and \emph{QuantumDifferentialParticleOptizerWithElitism} both have very good performance in higher dimensions for the first $2000$ function evaluations (even better than CMA-best), but around $10\,000$ function evaluations CMA-ES outperforms the other algorithms clearly. 
It is also interesting to observe that the optimized 
(for $d=5$) CMA-best performs less well than a CMA-ES with default hyper-parameters in $10d$ and $20d$.
Just like the \LLMEA\ proposed algorithms, CMA-ES best was optimized on $5d$ and with a budget of $10\,000$ evaluations. 
When looking at convergence curves per function, we note that CMA-ES outperforms mainly on the very hard functions in the benchmark ($f_{23}$ and $f_{24}$). 
Detailed convergence curves per BBOB function in $5d$, $10d$ and $20d$ are available in the supplemental material \cite{anonymous_2024_11358117}.

\reply{3}{5}{We would like to emphasize that the goal of our research is to show that, for a \emph{specific} setting (here: BBOB in $5d$), an LLM can generate a superior algorithm automatically.
This algorithm is not expected to scale to high-dimensional problems, as it was generated for the specific case of $5d$ problems. 
For this reason, we do not go beyond $d=20$ for further evaluation of the generated algorithms.}

\begin{figure*}[!ht]
    \centering
    \includegraphics[width=0.48\textwidth, trim=0mm 35mm 0mm 0mm,clip]{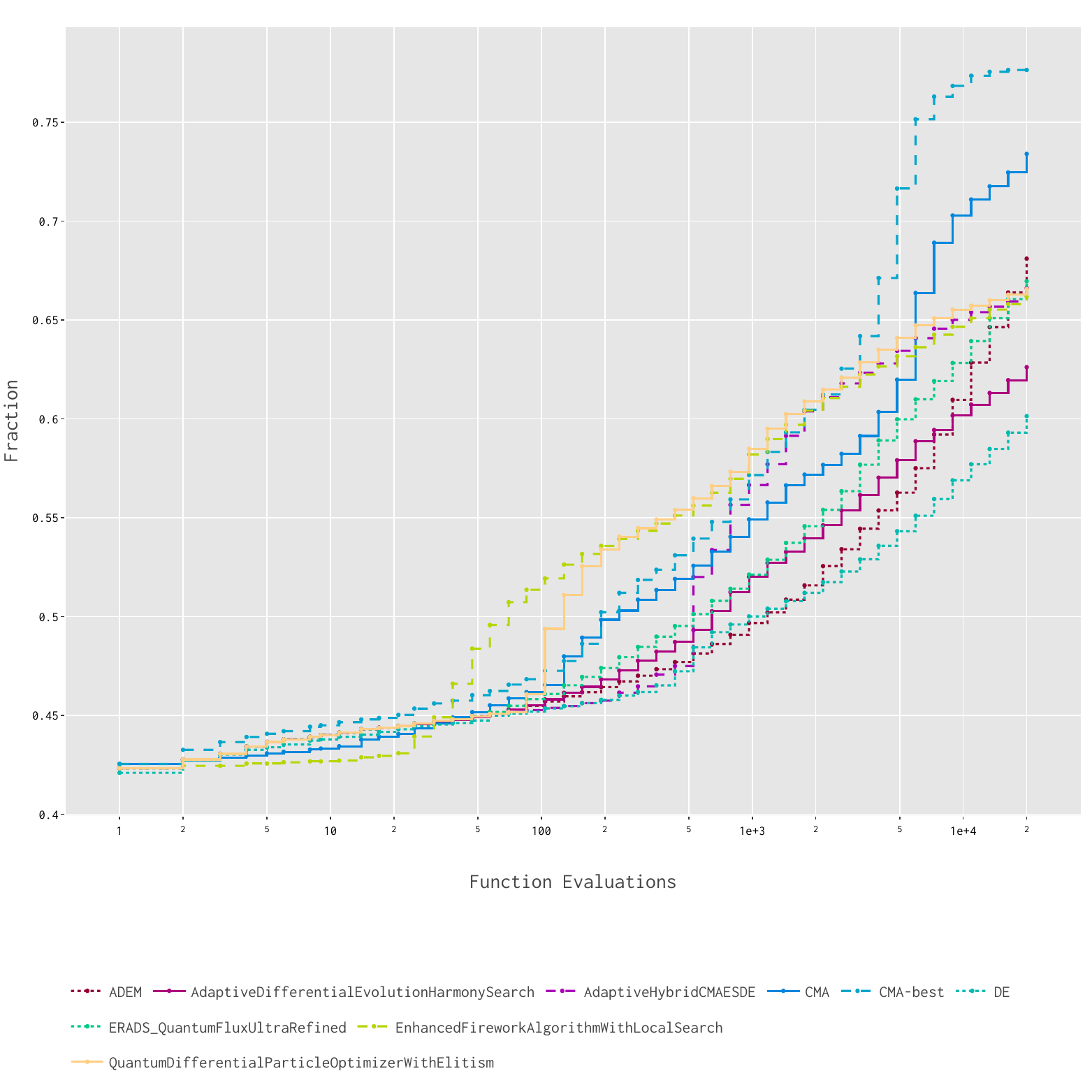}
    \includegraphics[width=0.48\textwidth, trim=0mm 35mm 0mm 0mm,clip]{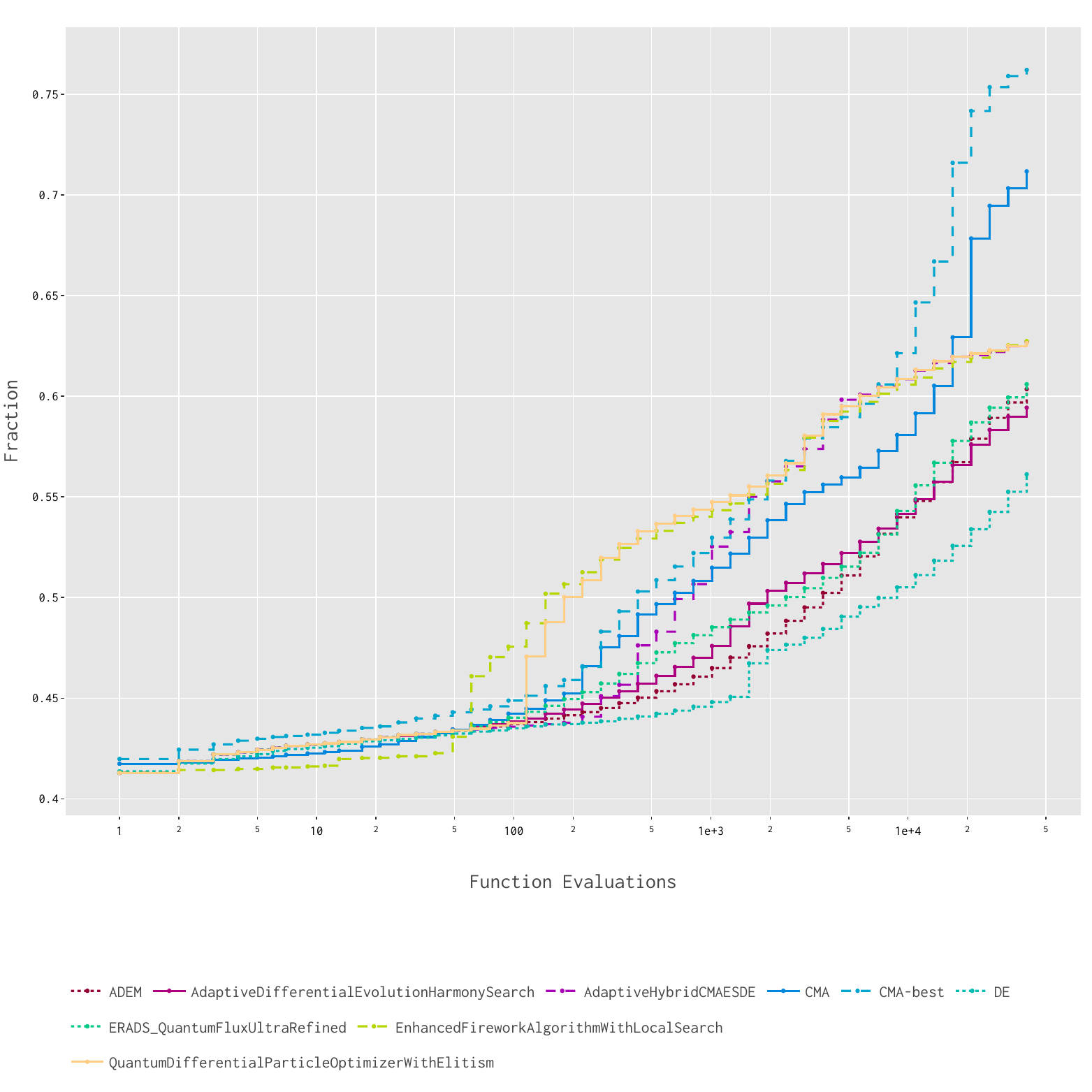}
    \includegraphics[width=0.8\textwidth, trim=10mm 0mm 0mm 235mm,clip]{plots/EAF_5d_best6.pdf}
    \caption{The empirical attainment function (EAF) estimates the percentage of runs that attain an arbitrary target value not later than a given runtime. EAF for the best algorithm per configuration and the baselines; CMA-ES, CMA-best and DE, averaged over all 24 BBOB functions in  $10d$ (left plot) and $20d$ (right plot), respectively.}\label{fig:EAFv2} 
\end{figure*}

\section{Conclusions and Outlook}\label{sec:conclusions}

This paper introduced \LLMEA, a novel framework leveraging Large Language Models (LLMs) for the automatic generation and optimization of metaheuristic algorithms. 
Our approach automates the evolution of algorithm design, enabling efficient exploration and optimization within a computationally feasible framework. 
Our findings demonstrate that \LLMEA\ can effectively generate high-performing algorithms that rival and sometimes surpass existing state-of-the-art techniques.

The \LLMEA\ framework proved capable of generating and evolving algorithms that perform comparably to traditional state-of-the-art metaheuristics, demonstrating the potential of LLMs to 
innovate within the algorithmic design space effectively.
Algorithms evolved through our framework successfully compete against established metaheuristics, highlighting the practical applicability of using LLMs for automated construction of optimization heuristics.
Since LLMs have been trained on an exceedingly large code base, including the currently available metaheuristics, we can explain their observed strong performance by interpreting them as a  \emph{universal modular framework for algorithm construction} (see also Section  \ref{sec:introduction})
that have access to a huge number of ``modules'' that can be combined to form new algorithms. In addition, we observed that the LLM is able to both fine-tune algorithm parameters, as well as introduce new logic such as different mutation and crossover strategies in the generated algorithms.

\clari{Considering how much work goes into the manual design of ``nature-inspired heuristics'', which are often not novel 
\cite{AranhaCCDRSSS22,Camacho-Villalon19,camacho2022analysis,StuetzleGreyWolf} and often cannot compete with random search \cite{10.1145/3638529.3654122}, we argue that an automated design approach for specific application domains will likely be the method of choice from now on.}

\subsubsection*{Challenges and limitations} 
The proposed \LLMEA\ framework presents several limitations and challenges that can be addressed for further advancement. One significant challenge lies in the dependency on the quality and structure of the prompts used to guide the LLM, which can introduce biases or limit the diversity of the generated algorithms. Additionally, the execution reliability of the dynamically generated code can be problematic, as errors during runtime can affect the evaluation of algorithm performance. Moreover, the computational cost associated with training and querying large language models may pose scalability issues, particularly for extensive or multi-objective optimization problems. Addressing these challenges would enhance the robustness and applicability of the \LLMEA\ framework across different domains and more complex optimization scenarios.

\subsubsection*{Outlook}
The promising results of the \LLMEA\ framework pave the way for several exciting directions for future research:
\begin{itemize}
    \item Future work could explore the expansion of the \LLMEA\ framework to support a broader range of evolutionary strategies. While the proposed framework focuses on an \ES\ approach, where we have one parent and generate one solution at a time, it is possible to generalize the framework to a \MLES\ \clari{(i.e., an algorithm with parent population size $\mu$ and offspring population size $\lambda \gg \mu$), as in population-based evolutionary algorithms \cite{back1997handbook}}, keeping a larger population and generating multiple individuals. This would translate into generating multiple mutations \clari{and recombinations} with LLMs by leveraging multiple random seeds \clari{and different temperature values} in the generation process.
    \item For generating diverse and innovative algorithm candidates, a population-based \MLES\ could use different LLMs in parallel and control the \emph{temperature parameter} of the LLMs to behave more explorative in the beginning of the search \cite{peeperkorn2024temperature}.
    \item Applying the \LLMEA\ methodology to other \clari{algorithm classes for continuous optimization problems, such as Bayesian and more general surrogate-assisted algorithms, could further demonstrate the versatility of \LLMEA.}
\end{itemize}

By continuing to develop and refine these approaches, we anticipate that the integration of LLMs in algorithmic design will significantly advance the field of evolutionary computation, leading to more intelligent, adaptable, and efficient systems.









\section*{Declarations}


\textbf{Disclosure of Interests.} The authors have no competing interests to declare that are relevant to the content of this article.

\medskip

\noindent \textbf{Reproducibility statement.}
We provide an open-source documented implementation of our package at \cite{anonymous_2024_11358117}.
Intermediate results, generated algorithms and BBOB evaluation results are available as well in subdirectories of the repository.



\FloatBarrier

\bibliographystyle{IEEEtran}
\bibliography{biblio.bib}

\begin{thebibliography}{10}
\providecommand{\url}[1]{#1}
\csname url@samestyle\endcsname
\providecommand{\newblock}{\relax}
\providecommand{\bibinfo}[2]{#2}
\providecommand{\BIBentrySTDinterwordspacing}{\spaceskip=0pt\relax}
\providecommand{\BIBentryALTinterwordstretchfactor}{4}
\providecommand{\BIBentryALTinterwordspacing}{\spaceskip=\fontdimen2\font plus
\BIBentryALTinterwordstretchfactor\fontdimen3\font minus \fontdimen4\font\relax}
\providecommand{\BIBforeignlanguage}[2]{{%
\expandafter\ifx\csname l@#1\endcsname\relax
\typeout{** WARNING: IEEEtran.bst: No hyphenation pattern has been}%
\typeout{** loaded for the language `#1'. Using the pattern for}%
\typeout{** the default language instead.}%
\else
\language=\csname l@#1\endcsname
\fi
#2}}
\providecommand{\BIBdecl}{\relax}
\BIBdecl

\bibitem{eiben2015introduction}
A.~E. Eiben and J.~E. Smith, \emph{Introduction to evolutionary computing}.\hskip 1em plus 0.5em minus 0.4em\relax Springer, 2015.

\bibitem{back1997handbook}
T.~B{\"a}ck, D.~B. Fogel, and Z.~Michalewicz, ``Handbook of evolutionary computation,'' \emph{Release}, vol.~97, no.~1, p.~B1, 1997.

\bibitem{KESwarmIntelligence}
J.~Kennedy and R.~C. Eberhart, \emph{Swarm Intelligence}.\hskip 1em plus 0.5em minus 0.4em\relax Morgan Kaufmann, 2001.

\bibitem{Dorigo2004ACO}
M.~Dorigo and T.~St\"utzle, \emph{Ant Colony Optimization}.\hskip 1em plus 0.5em minus 0.4em\relax MIT Press, 2004.

\bibitem{greiner2022evolutionary}
D.~J. Greiner~S{\'a}nchez, A.~Gaspar-Cunha, J.~D. Hern{\'a}ndez~Sosa, E.~Minisci, and A.~Zamuda, \emph{Evolutionary algorithms in engineering design optimization}.\hskip 1em plus 0.5em minus 0.4em\relax MDPI, 2022.

\bibitem{doi:10.1142/9781848166820_0006}
\BIBentryALTinterwordspacing
H.~Iba and N.~Noman, \emph{Real-world Applications of Evolutionary Algorithms}.\hskip 1em plus 0.5em minus 0.4em\relax Imperial College Press, UK, 2011, pp. 211--262. [Online]. Available: \url{https://www.worldscientific.com/doi/abs/10.1142/9781848166820_0006}
\BIBentrySTDinterwordspacing

\bibitem{chiong2012variants}
R.~Chiong, T.~Weise, and Z.~Michalewicz, \emph{Variants of evolutionary algorithms for real-world applications}.\hskip 1em plus 0.5em minus 0.4em\relax Springer, 2012, vol.~2.

\bibitem{ma2023performance}
Z.~Ma, G.~Wu, P.~N. Suganthan, A.~Song, and Q.~Luo, ``Performance assessment and exhaustive listing of 500+ nature-inspired metaheuristic algorithms,'' \emph{Swarm and Evolutionary Computation}, vol.~77, p. 101248, 2023.

\bibitem{del2021more}
J.~Del~Ser, E.~Osaba, A.~D. Martinez, M.~N. Bilbao, J.~Poyatos, D.~Molina, and F.~Herrera, ``More is not always better: insights from a massive comparison of meta-heuristic algorithms over real-parameter optimization problems,'' in \emph{2021 IEEE Symposium Series on Computational Intelligence (SSCI)}.\hskip 1em plus 0.5em minus 0.4em\relax IEEE, 2021, pp. 1--7.

\bibitem{10.1145/3638529.3654122}
\BIBentryALTinterwordspacing
D.~Vermetten, C.~Doerr, H.~Wang, A.~V. Kononova, and T.~B\"{a}ck, ``Large-scale benchmarking of metaphor-based optimization heuristics,'' in \emph{Proceedings of the Genetic and Evolutionary Computation Conference}, ser. GECCO '24.\hskip 1em plus 0.5em minus 0.4em\relax New York, NY, USA: Association for Computing Machinery, 2024, p. 41–49. [Online]. Available: \url{https://doi.org/10.1145/3638529.3654122}
\BIBentrySTDinterwordspacing

\bibitem{modES16}
S.~van Rijn, H.~Wang, M.~van Leeuwen, and T.~Bäck, ``Evolving the structure of evolution strategies,'' in \emph{2016 IEEE Symposium Series on Computational Intelligence (SSCI)}, 2016, pp. 1--8.

\bibitem{modCMA23}
\BIBentryALTinterwordspacing
D.~Vermetten, M.~L\'{o}pez-Ib\'{a}\~{n}ez, O.~Mersmann, R.~Allmendinger, and A.~V. Kononova, ``Analysis of modular {CMA-ES} on strict box-constrained problems in the {SBOX-COST} benchmarking suite,'' in \emph{Proceedings of the Companion Conference on Genetic and Evolutionary Computation}, ser. GECCO '23 Companion.\hskip 1em plus 0.5em minus 0.4em\relax New York, NY, USA: Association for Computing Machinery, 2023, p. 2346–2353. [Online]. Available: \url{https://doi.org/10.1145/3583133.3596419}
\BIBentrySTDinterwordspacing

\bibitem{modDE23}
\BIBentryALTinterwordspacing
D.~Vermetten, F.~Caraffini, A.~V. Kononova, and T.~B\"{a}ck, ``Modular differential evolution,'' in \emph{Proceedings of the Genetic and Evolutionary Computation Conference}, ser. GECCO '23.\hskip 1em plus 0.5em minus 0.4em\relax New York, NY, USA: Association for Computing Machinery, 2023, p. 864–872. [Online]. Available: \url{https://doi.org/10.1145/3583131.3590417}
\BIBentrySTDinterwordspacing

\bibitem{PSOX22}
C.~L. Camacho-Villalón, M.~Dorigo, and T.~Stützle, ``Pso-x: A component-based framework for the automatic design of particle swarm optimization algorithms,'' \emph{IEEE Transactions on Evolutionary Computation}, vol.~26, no.~3, pp. 402--416, 2022.

\bibitem{Design23}
\BIBentryALTinterwordspacing
C.~L. Camacho-Villalón, T.~Stützle, and M.~Dorigo, ``Designing new metaheuristics: Manual versus automatic approaches,'' \emph{Intelligent Computing}, vol.~2, p. 0048, 2023. [Online]. Available: \url{https://spj.science.org/doi/abs/10.34133/icomputing.0048}
\BIBentrySTDinterwordspacing

\bibitem{denobel2022iohexperimenterbenchmarkingplatformiterative}
\BIBentryALTinterwordspacing
J.~de~Nobel, F.~Ye, D.~Vermetten, H.~Wang, C.~Doerr, and T.~Bäck, ``Iohexperimenter: Benchmarking platform for iterative optimization heuristics,'' 2022. [Online]. Available: \url{https://arxiv.org/abs/2111.04077}
\BIBentrySTDinterwordspacing

\bibitem{10.1162/evco_a_00342}
\BIBentryALTinterwordspacing
------, ``{{IOH}experimenter: Benchmarking Platform for Iterative Optimization Heuristics},'' \emph{Evolutionary Computation}, pp. 1--6, 02 2024. [Online]. Available: \url{https://doi.org/10.1162/evco\_a\_00342}
\BIBentrySTDinterwordspacing

\bibitem{DBLP:conf/cec/NeumannNQDNVAYWB23}
\BIBentryALTinterwordspacing
F.~Neumann, A.~Neumann, C.~Qian, A.~V. Do, J.~de~Nobel, D.~Vermetten, S.~S. Ahouei, F.~Ye, H.~Wang, and T.~B{\"{a}}ck, ``Benchmarking algorithms for submodular optimization problems using {{IOH}}profiler,'' in \emph{{IEEE} Congress on Evolutionary Computation, {CEC} 2023, Chicago, IL, USA, July 1-5, 2023}.\hskip 1em plus 0.5em minus 0.4em\relax {IEEE}, 2023, pp. 1--9. [Online]. Available: \url{https://doi.org/10.1109/CEC53210.2023.10254181}
\BIBentrySTDinterwordspacing

\bibitem{DBLP:journals/asc/DoerrYHWSB20}
\BIBentryALTinterwordspacing
C.~Doerr, F.~Ye, N.~Horesh, H.~Wang, O.~M. Shir, and T.~B{\"{a}}ck, ``Benchmarking discrete optimization heuristics with {{IOH}}profiler,'' \emph{Appl. Soft Comput.}, vol.~88, p. 106027, 2020. [Online]. Available: \url{https://doi.org/10.1016/j.asoc.2019.106027}
\BIBentrySTDinterwordspacing

\bibitem{DBLP:journals/corr/abs-1810-05281}
\BIBentryALTinterwordspacing
C.~Doerr, H.~Wang, F.~Ye, S.~van Rijn, and T.~B{\"{a}}ck, ``{{IOH}}profiler: {A} benchmarking and profiling tool for iterative optimization heuristics,'' \emph{CoRR}, vol. abs/1810.05281, 2018. [Online]. Available: \url{http://arxiv.org/abs/1810.05281}
\BIBentrySTDinterwordspacing

\bibitem{DBLP:conf/gecco/DoerrWVBNY23}
\BIBentryALTinterwordspacing
C.~Doerr, H.~Wang, D.~Vermetten, T.~B{\"{a}}ck, J.~de~Nobel, and F.~Ye, ``Benchmarking and analyzing iterative optimization heuristics with {{IOH}}profiler,'' in \emph{Companion Proceedings of the Conference on Genetic and Evolutionary Computation, {GECCO} 2023, Companion Volume, Lisbon, Portugal, July 15-19, 2023}, S.~Silva and L.~Paquete, Eds.\hskip 1em plus 0.5em minus 0.4em\relax {ACM}, 2023, pp. 938--945. [Online]. Available: \url{https://doi.org/10.1145/3583133.3595057}
\BIBentrySTDinterwordspacing

\bibitem{DBLP:journals/telo/WangVYDB22}
\BIBentryALTinterwordspacing
H.~Wang, D.~Vermetten, F.~Ye, C.~Doerr, and T.~B{\"{a}}ck, ``{{IOH}}analyzer: Detailed performance analyses for iterative optimization heuristics,'' \emph{{ACM} Trans. Evol. Learn. Optim.}, vol.~2, no.~1, pp. 3:1--3:29, 2022. [Online]. Available: \url{https://doi.org/10.1145/3510426}
\BIBentrySTDinterwordspacing

\bibitem{hansen2010black}
N.~Hansen and R.~Ros, ``Black-box optimization benchmarking of {NEWUOA} compared to {BIPOP-CMA-ES:} on the {BBOB} noiseless testbed,'' in \emph{Proceedings of the 12th annual conference companion on Genetic and evolutionary computation}, 2010, pp. 1519--1526.

\bibitem{eafecdf}
M.~López-Ibáñez, D.~Vermetten, J.~Dreo, and C.~Doerr, ``Using the empirical attainment function for analyzing single-objective black-box optimization algorithms,'' 2024, arXiv:2404.02031.

\bibitem{guo2024connecting}
Q.~Guo, R.~Wang, J.~Guo, B.~Li, K.~Song, X.~Tan, G.~Liu, J.~Bian, and Y.~Yang, ``Connecting large language models with evolutionary algorithms yields powerful prompt optimizers,'' 2024, arXiv:2309.08532.

\bibitem{rios2023large}
T.~Rios, S.~Menzel, and B.~Sendhoff, ``Large language and text-to-3d models for engineering design optimization,'' 2023, arXiv:2307.01230.

\bibitem{zhou2022large}
\BIBentryALTinterwordspacing
Y.~Zhou, A.~I. Muresanu, Z.~Han, K.~Paster, S.~Pitis, H.~Chan, and J.~Ba, ``Large language models are human-level prompt engineers,'' in \emph{NeurIPS 2022 Foundation Models for Decision Making Workshop}, 2022. [Online]. Available: \url{https://openreview.net/forum?id=YdqwNaCLCx}
\BIBentrySTDinterwordspacing

\bibitem{honovich-etal-2023-instruction}
\BIBentryALTinterwordspacing
O.~Honovich, U.~Shaham, S.~R. Bowman, and O.~Levy, ``Instruction induction: From few examples to natural language task descriptions,'' in \emph{Proceedings of the 61st Annual Meeting of the Association for Computational Linguistics (Volume 1: Long Papers)}, A.~Rogers, J.~Boyd-Graber, and N.~Okazaki, Eds.\hskip 1em plus 0.5em minus 0.4em\relax Toronto, Canada: Association for Computational Linguistics, Jul. 2023, pp. 1935--1952. [Online]. Available: \url{https://aclanthology.org/2023.acl-long.108}
\BIBentrySTDinterwordspacing

\bibitem{Kojima2022LargeLM}
\BIBentryALTinterwordspacing
T.~Kojima, S.~S. Gu, M.~Reid, Y.~Matsuo, and Y.~Iwasawa, ``Large language models are zero-shot reasoners,'' \emph{ArXiv}, vol. abs/2205.11916, 2022. [Online]. Available: \url{https://api.semanticscholar.org/CorpusID:249017743}
\BIBentrySTDinterwordspacing

\bibitem{lange2024large}
R.~T. Lange, Y.~Tian, and Y.~Tang, ``Large language models as evolution strategies,'' 2024, arXiv:2402.18381.

\bibitem{hansen2009real}
N.~Hansen, S.~Finck, R.~Ros, and A.~Auger, ``Real-parameter black-box optimization benchmarking 2009: Noiseless functions definitions,'' INRIA, Tech. Rep. RR6829, 2009.

\bibitem{luo2024incontext}
M.~Luo, X.~Xu, Y.~Liu, P.~Pasupat, and M.~Kazemi, ``In-context learning with retrieved demonstrations for language models: A survey,'' 2024, arXiv:2401.11624.

\bibitem{dong2023survey}
Q.~Dong, L.~Li, D.~Dai, C.~Zheng, Z.~Wu, B.~Chang, X.~Sun, J.~Xu, L.~Li, and Z.~Sui, ``A survey on in-context learning,'' 2023, arXiv:2301.00234.

\bibitem{FunSearch2024}
B.~Romera-Paredes, M.~Barekatain, A.~Novikov, M.~Balog, M.~P. Kumar, E.~Dupont, F.~J. Ruiz, J.~S. Ellenberg, P.~Wang, O.~Fawzi, P.~Kohli, and A.~Fawzi, ``{Mathematical discoveries from program search with large language models},'' \emph{Nature}, vol. 625, pp. 468--475, 01 2024.

\bibitem{liu2023algorithm}
F.~Liu, X.~Tong, M.~Yuan, and Q.~Zhang, ``Algorithm evolution using large language model,'' 2023, arXiv:2311.15249.

\bibitem{fei2024eoh}
\BIBentryALTinterwordspacing
L.~Fei, X.~Tong, M.~Yuan, X.~Lin, F.~Luo, Z.~Wang, Z.~Lu, and Q.~Zhang, ``Evolution of heuristics: Towards efficient automatic algorithm design using large language model,'' in \emph{International Conference on Machine Learning (ICML)}, 2024. [Online]. Available: \url{https://arxiv.org/abs/2401.02051}
\BIBentrySTDinterwordspacing

\bibitem{zelikman2024selftaught}
E.~Zelikman, E.~Lorch, L.~Mackey, and A.~T. Kalai, ``Self-taught optimizer ({STOP}): Recursively self-improving code generation,'' 2024, arXiv:2310.02304.

\bibitem{PluhacekLLMmetaheuristics2023}
\BIBentryALTinterwordspacing
M.~Pluhacek, A.~Kazikova, T.~Kadavy, A.~Viktorin, and R.~Senkerik, ``Leveraging large language models for the generation of novel metaheuristic optimization algorithms,'' in \emph{Proceedings of the Companion Conference on Genetic and Evolutionary Computation}, ser. GECCO '23 Companion.\hskip 1em plus 0.5em minus 0.4em\relax New York, NY, USA: Association for Computing Machinery, 2023, p. 1812–1820. [Online]. Available: \url{https://doi.org/10.1145/3583133.3596401}
\BIBentrySTDinterwordspacing

\bibitem{IOHexperimenter}
\BIBentryALTinterwordspacing
J.~de~Nobel, F.~Ye, D.~Vermetten, H.~Wang, C.~Doerr, and T.~B{\"{a}}ck, ``{IOH}experimenter: Benchmarking platform for iterative optimization heuristics,'' \emph{arXiv e-prints:2111.04077}, nov 2021. [Online]. Available: \url{https://arxiv.org/abs/2111.04077}
\BIBentrySTDinterwordspacing

\bibitem{anonymous_2024_11358117}
\BIBentryALTinterwordspacing
N.~van Stein, ``{LLaMEA},'' Jun. 2024, zenodo. [Online]. Available: \url{https://doi.org/10.5281/zenodo.11358116}
\BIBentrySTDinterwordspacing

\bibitem{cheng2018model}
R.~Cheng, C.~He, Y.~Jin, and X.~Yao, ``Model-based evolutionary algorithms: a short survey,'' \emph{Complex \& Intelligent Systems}, vol.~4, no.~4, pp. 283--292, 2018.

\bibitem{khaldi2023surrogate}
M.~I.~E. Khaldi and A.~Draa, ``Surrogate-assisted evolutionary optimisation: a novel blueprint and a state of the art survey,'' \emph{Evolutionary Intelligence}, pp. 1--31, 2023.

\bibitem{openai2022chatgpt35turbo}
OpenAI, ``Chatgpt-3.5-turbo,'' \url{https://platform.openai.com/docs/models/gpt-3-5-turbo}, 2022, version 0125, Accessed: 2024-05-01.

\bibitem{openai2023chatgpt4turbo}
------, ``Chatgpt-4-turbo,'' \url{https://platform.openai.com/docs/models/gpt-4-turbo-and-gpt-4}, 2023, version 2024-04-09, Accessed: 2024-05-01.

\bibitem{openai2023chatgpt4o}
------, ``Chatgpt-4o,'' \url{https://platform.openai.com/docs/models/gpt-4o}, 2023, version: 2024-05-13, Accessed: 2024-05-14.

\bibitem{vanstein2024explainable}
N.~van Stein, D.~Vermetten, A.~V. Kononova, and T.~Bäck, ``Explainable benchmarking for iterative optimization heuristics,'' 2024, arXiv:2401.17842.

\bibitem{hansen2022anytime}
N.~Hansen, A.~Auger, D.~Brockhoff, and T.~Tu{\v{s}}ar, ``Anytime performance assessment in blackbox optimization benchmarking,'' \emph{IEEE Transactions on Evolutionary Computation}, vol.~26, no.~6, pp. 1293--1305, 2022.

\bibitem{jaro89}
\BIBentryALTinterwordspacing
M.~A. Jaro, ``Advances in record-linkage methodology as applied to matching the 1985 census of {T}ampa, {F}lorida,'' \emph{Journal of the American Statistical Association}, vol.~84, no. 406, pp. 414--420, 1989. [Online]. Available: \url{https://www.tandfonline.com/doi/abs/10.1080/01621459.1989.10478785}
\BIBentrySTDinterwordspacing

\bibitem{brockhoff2012comparing}
D.~Brockhoff, A.~Auger, and N.~Hansen, ``Comparing mirrored mutations and active covariance matrix adaptation in the {IPOP-CMA-ES} on the noiseless {BBOB} testbed,'' in \emph{Proceedings of the 14th annual conference companion on Genetic and evolutionary computation}, 2012, pp. 297--304.

\bibitem{povsik2012benchmarking}
P.~Po{\v{s}}{\'\i}k and V.~Klem{\v{s}}, ``Benchmarking the differential evolution with adaptive encoding on noiseless functions,'' in \emph{Proceedings of the 14th annual conference companion on Genetic and evolutionary computation}, 2012, pp. 189--196.

\bibitem{modcma}
\BIBentryALTinterwordspacing
J.~de~Nobel, D.~Vermetten, H.~Wang, C.~Doerr, and T.~B{\"{a}}ck, ``Tuning as a means of assessing the benefits of new ideas in interplay with existing algorithmic modules,'' in \emph{Proc. of Genetic and Evolutionary Computation Conference (GECCO'21, Companion material)}.\hskip 1em plus 0.5em minus 0.4em\relax {ACM}, 2021, pp. 1375--1384. [Online]. Available: \url{https://doi.org/10.1145/3449726.3463167}
\BIBentrySTDinterwordspacing

\bibitem{huang2023survey}
L.~Huang, W.~Yu, W.~Ma, W.~Zhong, Z.~Feng, H.~Wang, Q.~Chen, W.~Peng, X.~Feng, B.~Qin, and T.~Liu, ``A survey on hallucination in large language models: Principles, taxonomy, challenges, and open questions,'' 2023, arXiv:2311.05232.

\bibitem{jade5208221}
J.~Zhang and A.~C. Sanderson, ``{JADE}: Adaptive differential evolution with optional external archive,'' \emph{IEEE Transactions on Evolutionary Computation}, vol.~13, no.~5, pp. 945--958, 2009.

\bibitem{XIA202133}
\BIBentryALTinterwordspacing
X.~Xia, L.~Tong, Y.~Zhang, X.~Xu, H.~Yang, L.~Gui, Y.~Li, and K.~Li, ``Nfdde: A novelty-hybrid-fitness driving differential evolution algorithm,'' \emph{Information Sciences}, vol. 579, pp. 33--54, 2021. [Online]. Available: \url{https://www.sciencedirect.com/science/article/pii/S0020025521007726}
\BIBentrySTDinterwordspacing

\bibitem{IOHanalyzer}
\BIBentryALTinterwordspacing
H.~Wang, D.~Vermetten, F.~Ye, C.~Doerr, and T.~B\"{a}ck, ``{IOH}analyzer: Detailed performance analyses for iterative optimization heuristics,'' \emph{ACM Transactions on Evolutionary Learning and Optimization}, vol.~2, no.~1, apr 2022. [Online]. Available: \url{https://doi.org/10.1145/3510426}
\BIBentrySTDinterwordspacing

\bibitem{AranhaCCDRSSS22}
\BIBentryALTinterwordspacing
C.~Aranha, C.~L. Camacho{-}Villal{\'{o}}n, F.~Campelo, M.~Dorigo, R.~Ruiz, M.~Sevaux, K.~S{\"{o}}rensen, and T.~St{\"{u}}tzle, ``Metaphor-based metaheuristics, a call for action: the elephant in the room,'' \emph{Swarm Intell.}, vol.~16, no.~1, pp. 1--6, 2022. [Online]. Available: \url{https://doi.org/10.1007/s11721-021-00202-9}
\BIBentrySTDinterwordspacing

\bibitem{Camacho-Villalon19}
\BIBentryALTinterwordspacing
C.~L. Camacho{-}Villal{\'{o}}n, M.~Dorigo, and T.~St{\"{u}}tzle, ``The intelligent water drops algorithm: why it cannot be considered a novel algorithm - {A} brief discussion on the use of metaphors in optimization,'' \emph{Swarm Intell.}, vol.~13, no. 3-4, pp. 173--192, 2019. [Online]. Available: \url{https://doi.org/10.1007/s11721-019-00165-y}
\BIBentrySTDinterwordspacing

\bibitem{camacho2022analysis}
C.~L. Camacho-Villal{\'o}n, M.~Dorigo, and T.~St{\"u}tzle, ``An analysis of why cuckoo search does not bring any novel ideas to optimization,'' \emph{Computers \& Operations Research}, vol. 142, p. 105747, 2022.

\bibitem{StuetzleGreyWolf}
\BIBentryALTinterwordspacing
C.~L. Camacho{-}Villal{\'{o}}n, T.~St{\"{u}}tzle, and M.~Dorigo, ``Grey wolf, firefly and bat algorithms: Three widespread algorithms that do not contain any novelty,'' in \emph{Proc. of Swarm Intelligence (ANTS)}, ser. LNCS, vol. 12421.\hskip 1em plus 0.5em minus 0.4em\relax Springer, 2020, pp. 121--133. [Online]. Available: \url{https://doi.org/10.1007/978-3-030-60376-2\_10}
\BIBentrySTDinterwordspacing

\bibitem{peeperkorn2024temperature}
M.~Peeperkorn, T.~Kouwenhoven, D.~Brown, and A.~Jordanous, ``Is temperature the creativity parameter of large language models?'' 2024, arXiv:2405.00492.

\bibitem{bbob_hansen2009_noiseless}
\BIBentryALTinterwordspacing
N.~Hansen, S.~Finck, R.~Ros, and A.~Auger, ``{Real-Parameter Black-Box Optimization Benchmarking 2009: Noiseless Functions Definitions},'' {INRIA}, Research Report RR-6829, 2009. [Online]. Available: \url{https://hal.inria.fr/inria-00362633}
\BIBentrySTDinterwordspacing

\bibitem{smac3}
\BIBentryALTinterwordspacing
M.~Lindauer, K.~Eggensperger, M.~Feurer, A.~Biedenkapp, D.~Deng, C.~Benjamins, T.~Ruhkopf, R.~Sass, and F.~Hutter, ``{SMAC3:} {A} versatile bayesian optimization package for hyperparameter optimization,'' \emph{J. Mach. Learn. Res.}, vol.~23, pp. 54:1--54:9, 2022. [Online]. Available: \url{https://jmlr.org/papers/v23/21-0888.html}
\BIBentrySTDinterwordspacing

\end{thebibliography}

\begin{IEEEbiography}[{\includegraphics[width=1in,height=1.25in,clip,keepaspectratio]{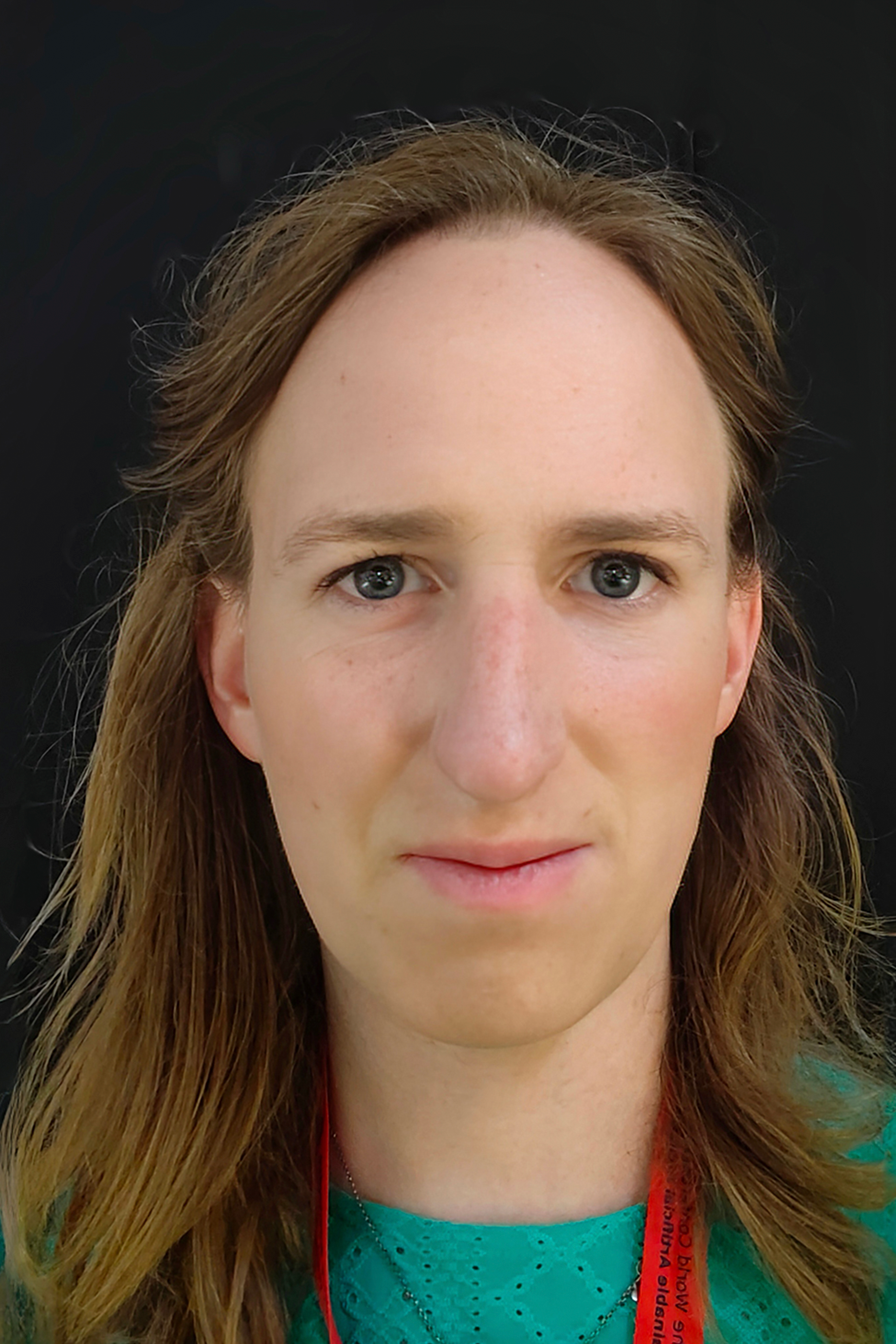}}]{Niki van Stein}
(M’22) received her Ph.D.~degree in Computer Science in 2018, from the Leiden Institute of Advanced Computer Science (LIACS), Leiden University, The Netherlands.
She is currently an Assistant Professor at LIACS, leading the XAI research group within the Natural Computing Cluster. Dr.~van Stein's research interests focus on integrating explainability into AI models to enhance transparency and trustworthiness, with applications in predictive maintenance, engineering design, and time-series analysis. She has a special interest in merging large language models with evolutionary computing techniques to innovate in algorithmic design.
\end{IEEEbiography}

\vspace{-33pt}

\begin{IEEEbiography}[{\includegraphics[width=1in,height=1.25in,clip,keepaspectratio]{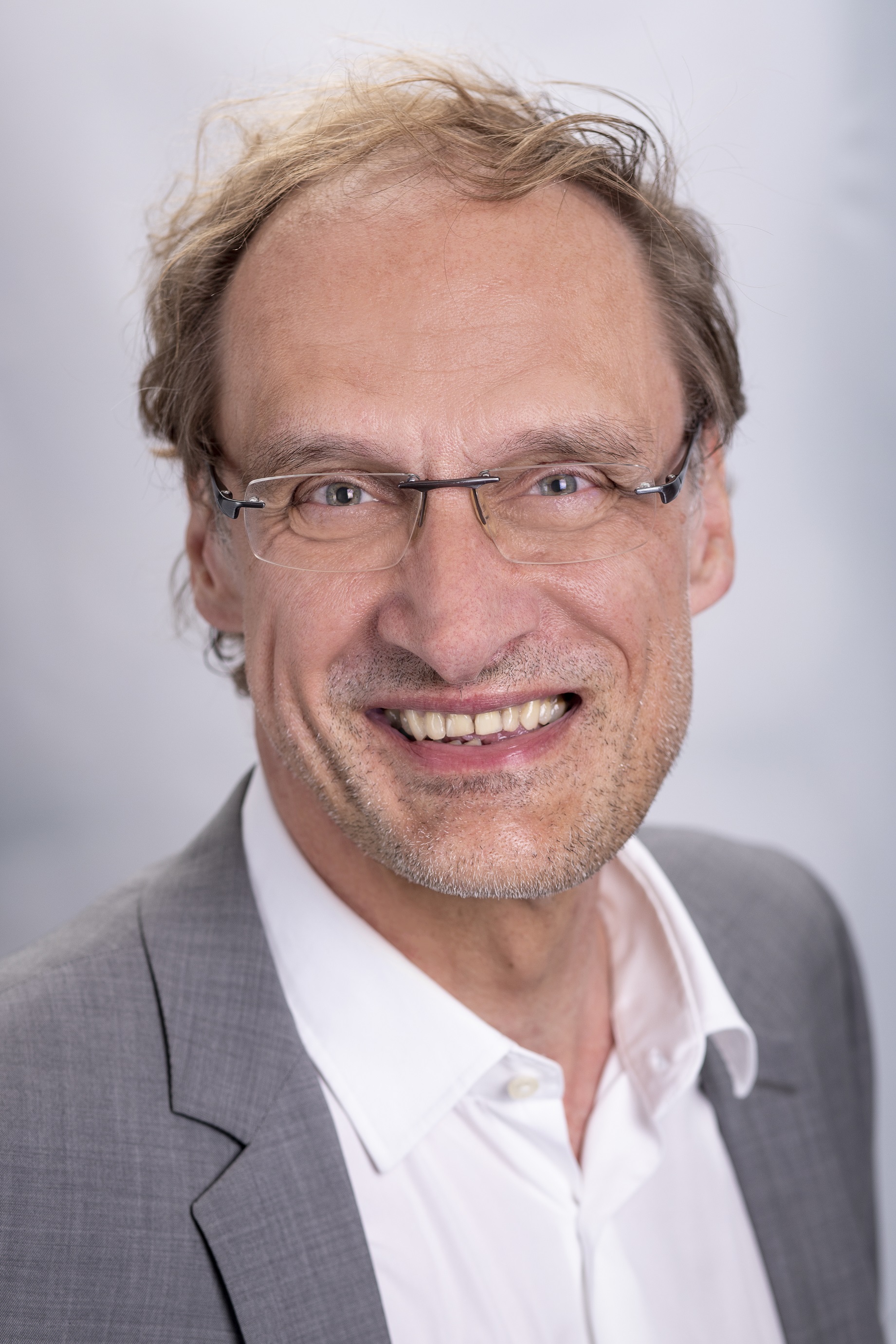}}]{Thomas B{\"a}ck} (M’95-SM’19-F’22) received the Diploma degree  (1990) and Ph.D.~degree (1994) in computer science from the University of Dortmund, Germany.
Since 2002, he is Full Professor of computer science at the Leiden Institute of Advanced Computer Science (LIACS), Leiden University, The Netherlands. He is author of Evolutionary Algorithms in Theory and Practice (OUP, 1996) and co-editor of the Handbook of Evolutionary Computation (CRC Press, 1997) and Handbook of Natural Computing (Springer, 2012). 
His research interests include evolutionary computation, machine learning and their real-world applications, especially in sustainable smart industry and health.

Prof.~B\"ack’s awards and honors include membership in the Royal Netherlands Academy of Arts and Sciences (KNAW, 2021) and Academia Europaea (2022), the IEEE Computational Intelligence Society Evolutionary Computation Pioneer Award (2015), ACM SIGEVO Outstanding Contribution Award (2023), and best Ph.D.~thesis award of the German society of Computer Science (GI, 1995).
\end{IEEEbiography}

\onecolumn

 {\appendices

\section{BBOB Function Groups}
\label{secA1}

In the experiments we heavily use the black-box optimization benchmarking suite (BBOB), which consists of $24$ different functions categorized in $5$ different function groups. In Table \ref{tab:groups}, each function, its name and its respective function group is given. For additional details see \cite{bbob_hansen2009_noiseless}. 
As it is common in BBOB, rotation and translation are applied on each of these functions to generate additional function instances. Using this wide range of different function landscapes and different random instances allows for developing (and, in this case, automatically discover) strong black-box optimization metaheuristics that perform well on a large range of problems.

\begin{table*}[!ht]
\centering
\caption{The 24 BBOB noiseless functions grouped in five function categories \label{tab:groups}}
\begin{tabular}{cll}
\toprule
\textbf{Group} & \textbf{Function ID} & \textbf{Function Name} \\
\midrule
1 & f1 & Sphere Function \\
  & f2 & Separable Ellipsoidal Function \\
  & f3 & Rastrigin Function \\
  & f4 & Büche-Rastrigin Function \\
  & f5 & Linear Slope \\
\midrule
2 & f6 & Attractive Sector Function \\
  & f7 & Step Ellipsoidal Function \\
  & f8 & Rosenbrock Function, original \\
  & f9 & Rosenbrock Function, rotated \\
\midrule
3 & f10 & Ellipsoidal Function \\
  & f11 & Discus Function \\
  & f12 & Bent Cigar Function \\
  & f13 & Sharp Ridge Function \\
  & f14 & Different Powers Function \\
\midrule
4 & f15 & Rastrigin Function \\
  & f16 & Weierstrass Function \\
  & f17 & Schaffer's F7 Function \\
  & f18 & Schaffer's F7 Function, moderately ill-conditioned \\
  & f19 & Composite Griewank-Rosenbrock Function F8F2 \\
\midrule
5 & f20 & Schwefel Function \\
  & f21 & Gallagher's Gaussian 101-me Peaks Function \\
  & f22 & Gallagher's Gaussian 21-hi Peaks Function \\
  & f23 & Katsuura Function \\
  & f24 & Lunacek bi-Rastrigin Function \\
\bottomrule
\end{tabular}
\end{table*}

\FloatBarrier
\section{Extended Results on BBOB}
\label{secA2}

In the following supplementary section we provided additional results of the BBOB experiments done with the best algorithms discovered by each LLM and strategy combination. This includes the strategy $1,1$ 
\emph{with details}. In this strategy we give the LLM additional information in the form of AOCC scores per BBOB function group, instead of one average AOCC score. Since this additional feedback did not result in better optimization performance, we only show these results in the supplemental section. In Figure \ref{fig:EAFv2} we can observe the performance of the $9$ algorithms that were the results of each LLaMEA configuration (three LLM versions and the three strategies (1+1, 1,1 and 1,1-with-details) 
against the three baselines, CMA-ES, optimized CMA-ES and DE (in $5d$, $10d$, and $20d$). In Figures \ref{fig:FCEd5}, \ref{fig:FCEd10} and \ref{fig:FCEd20}, the convergence curves (function values) per BBOB function of these algorithms are shown in respectively $5d$, $10d$ and $20d$.


\begin{figure*}[!ht]
    \centering
     \includegraphics[width=0.8\textwidth, trim=0mm 30mm 0mm 0mm,clip]{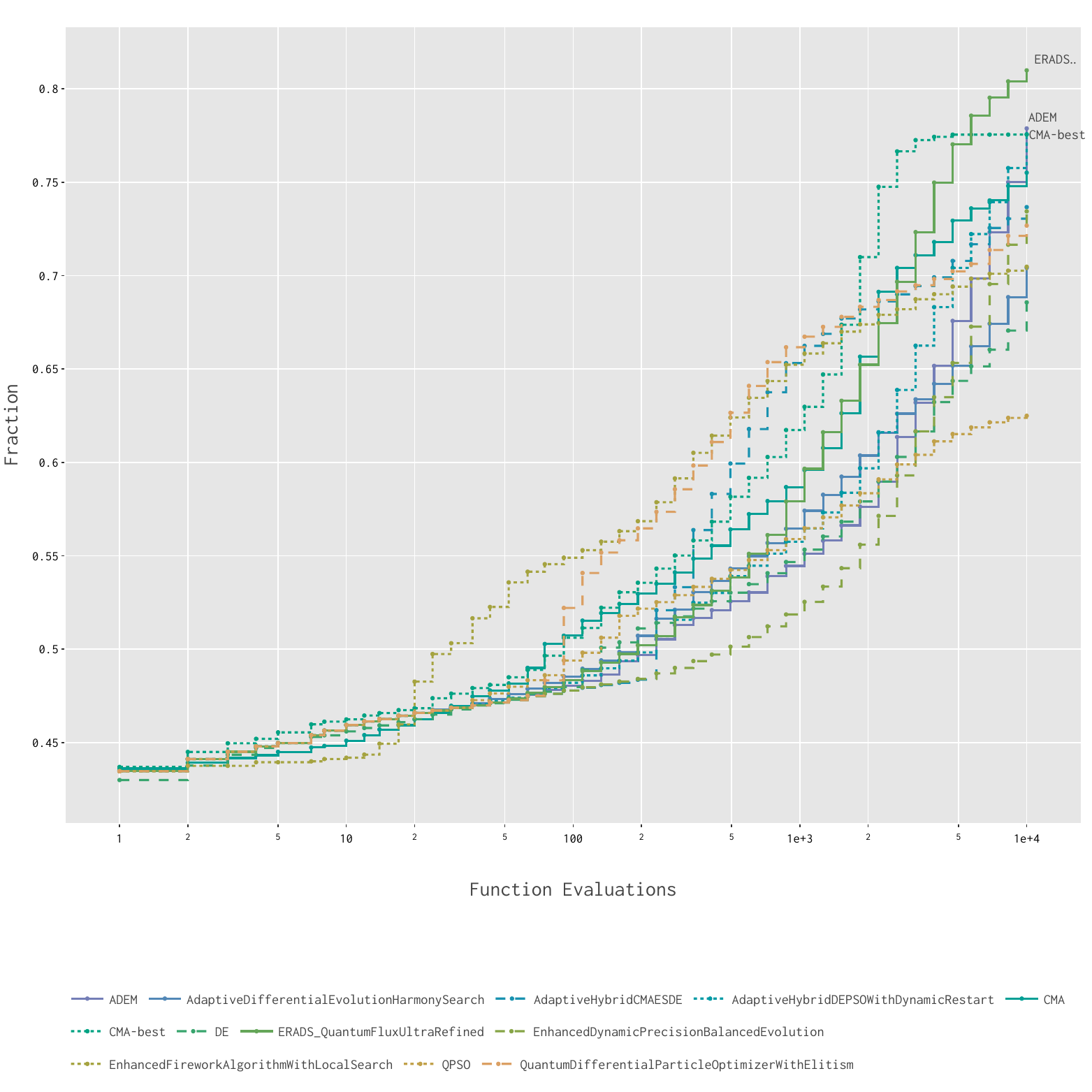}
    \includegraphics[width=0.4\textwidth, trim=0mm 30mm 0mm 0mm,clip]{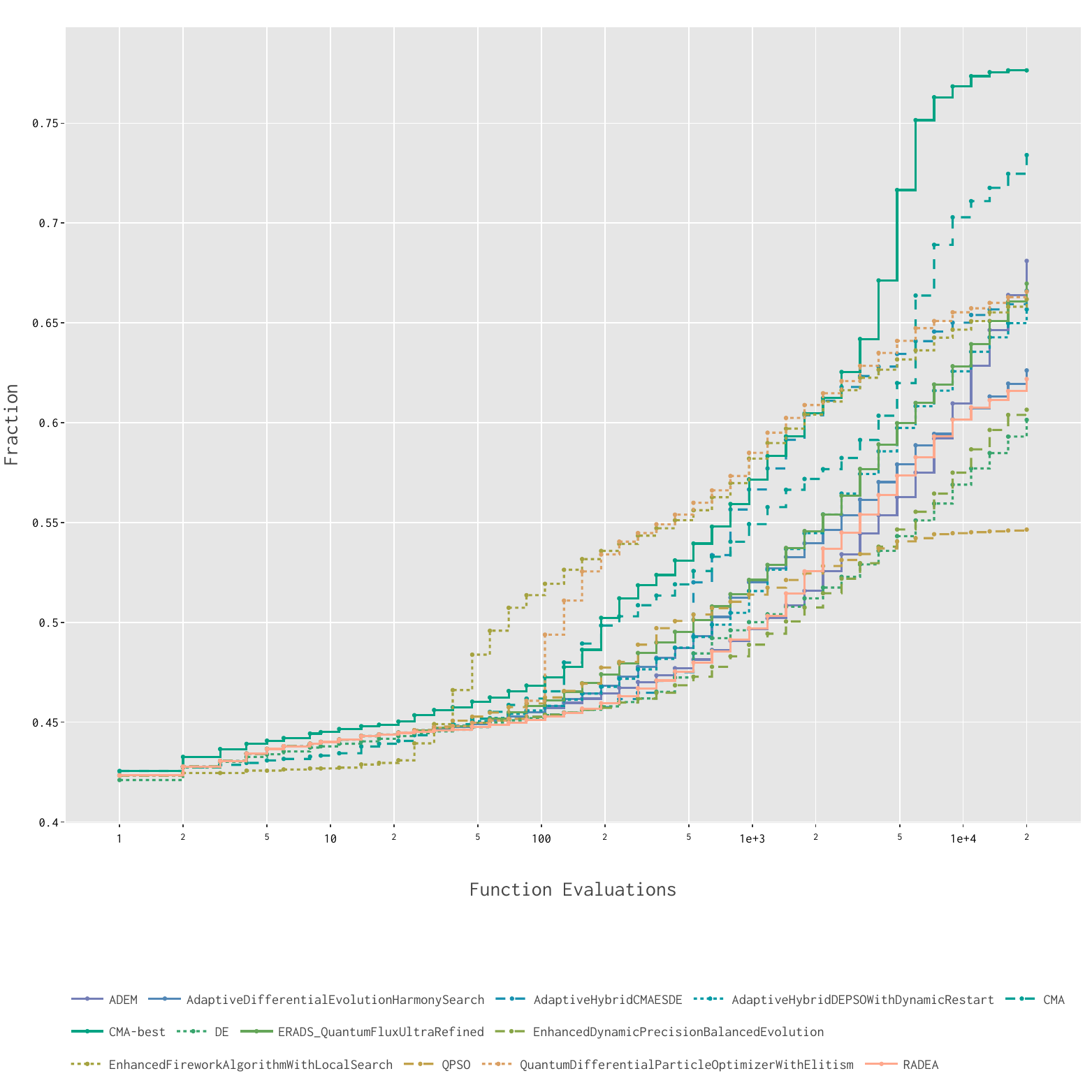}
    \includegraphics[width=0.4\textwidth, trim=0mm 30mm 0mm 0mm,clip]{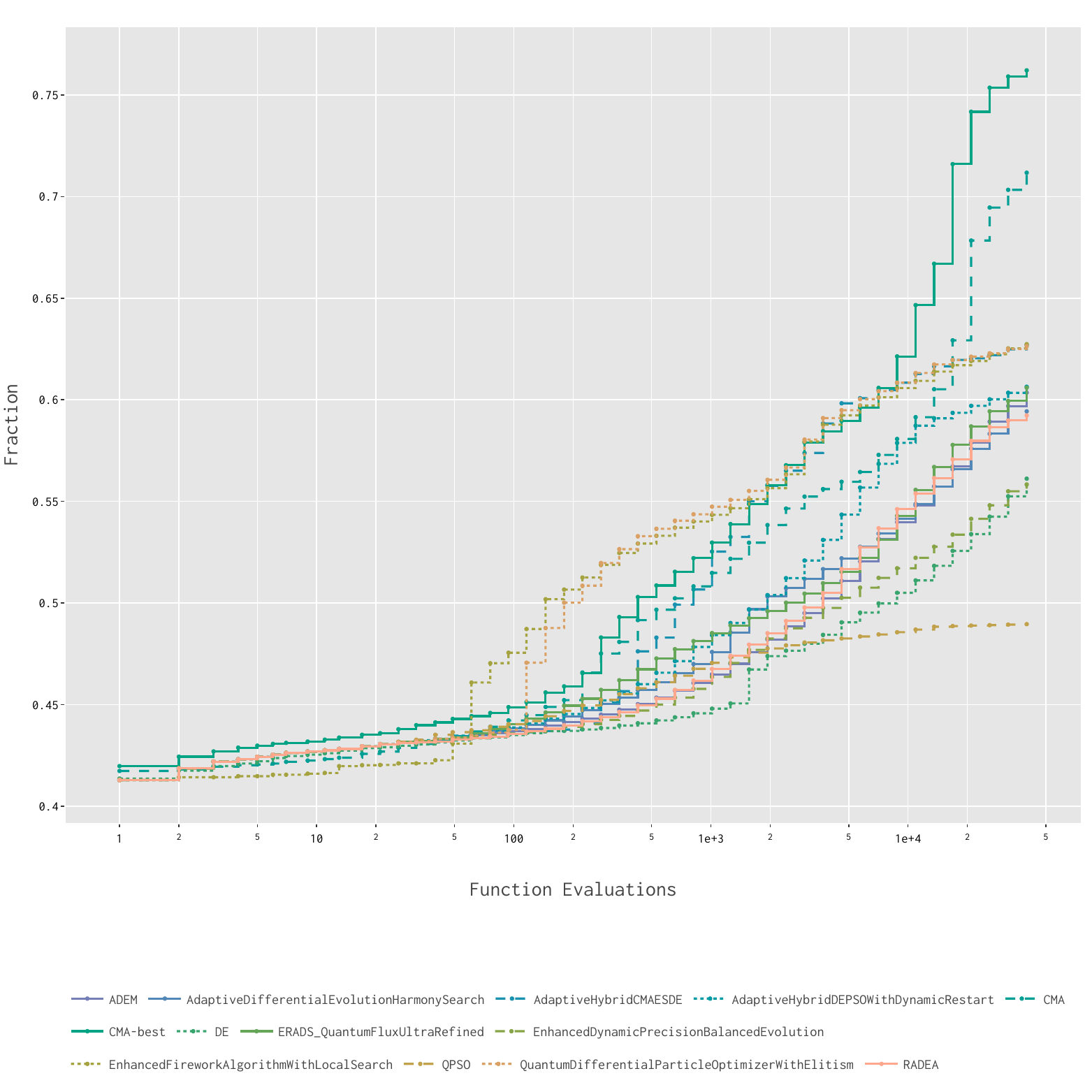}
    \includegraphics[width=\textwidth, trim=10mm 0mm 0mm 240mm,clip]{plots/EAF_5d_best27.pdf}
    \caption{The empirical attainment function (EAF) estimates the percentage of runs that attain an arbitrary target value not later than a given runtime. EAF for the best algorithm per configuration  and the baselines; CMA-ES, CMA-best and DE, averaged over all 24 BBOB functions in $5d$ (top plot),  $10d$ (left bottom plot) and $20d$ (right bottom plot), respectively.}\label{fig:EAFv2} 
\end{figure*}

\begin{figure*}[!ht]
    \centering
    \includegraphics[width=\textwidth, trim=8mm 0mm 0mm 0mm,clip]{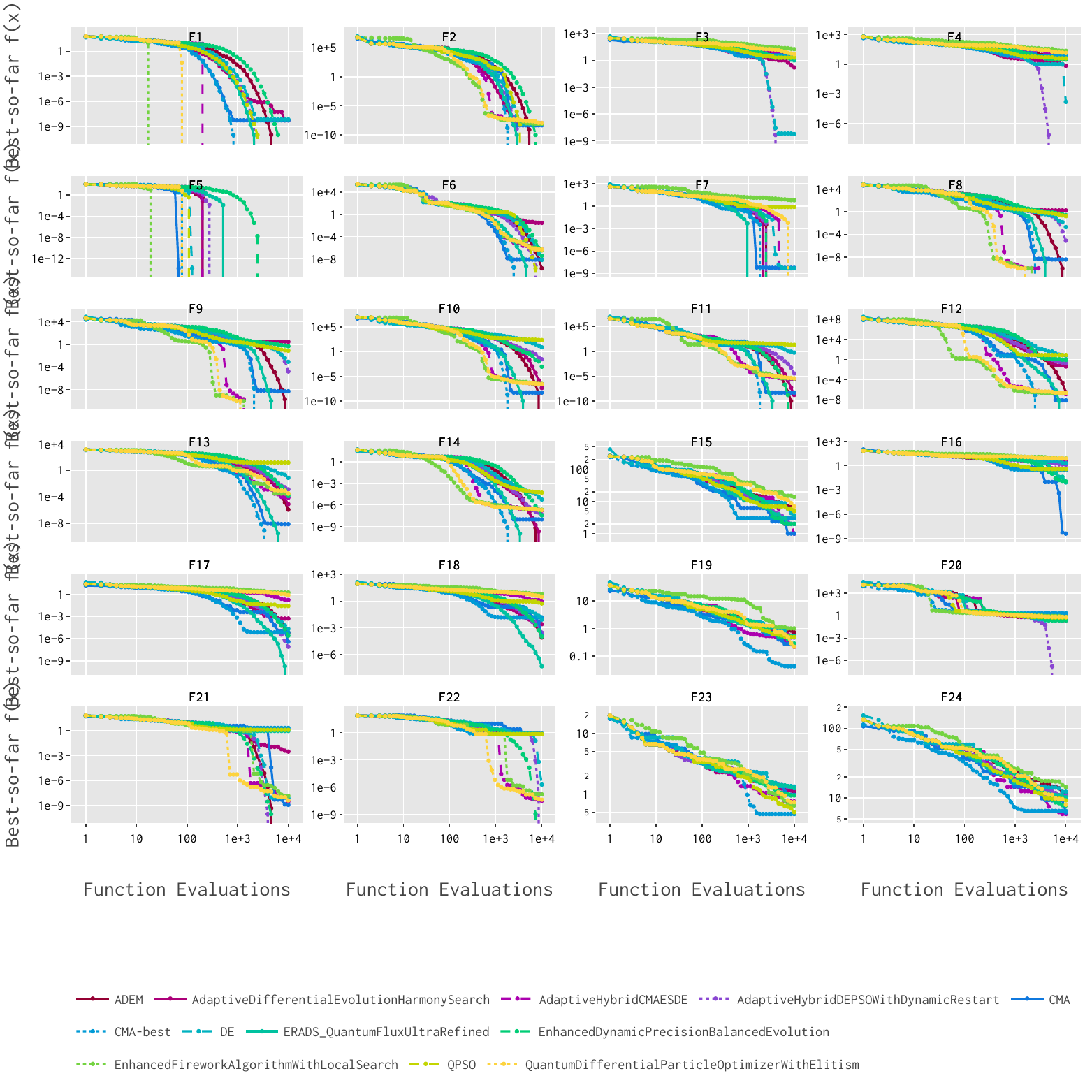}
    \caption{Median best-so-far function value ($y$-axis) over the $2\,000 \cdot d$ function evaluations ($x$-axis) per BBOB function for the best algorithm per configuration and the baselines; CMA-ES, DE and BIPOP-CMA-ES (for $d=5$).}\label{fig:FCEd5} 
\end{figure*}

\begin{figure*}[!ht]
    \centering
    \includegraphics[width=\textwidth, trim=8mm 0mm 0mm 0mm,clip]{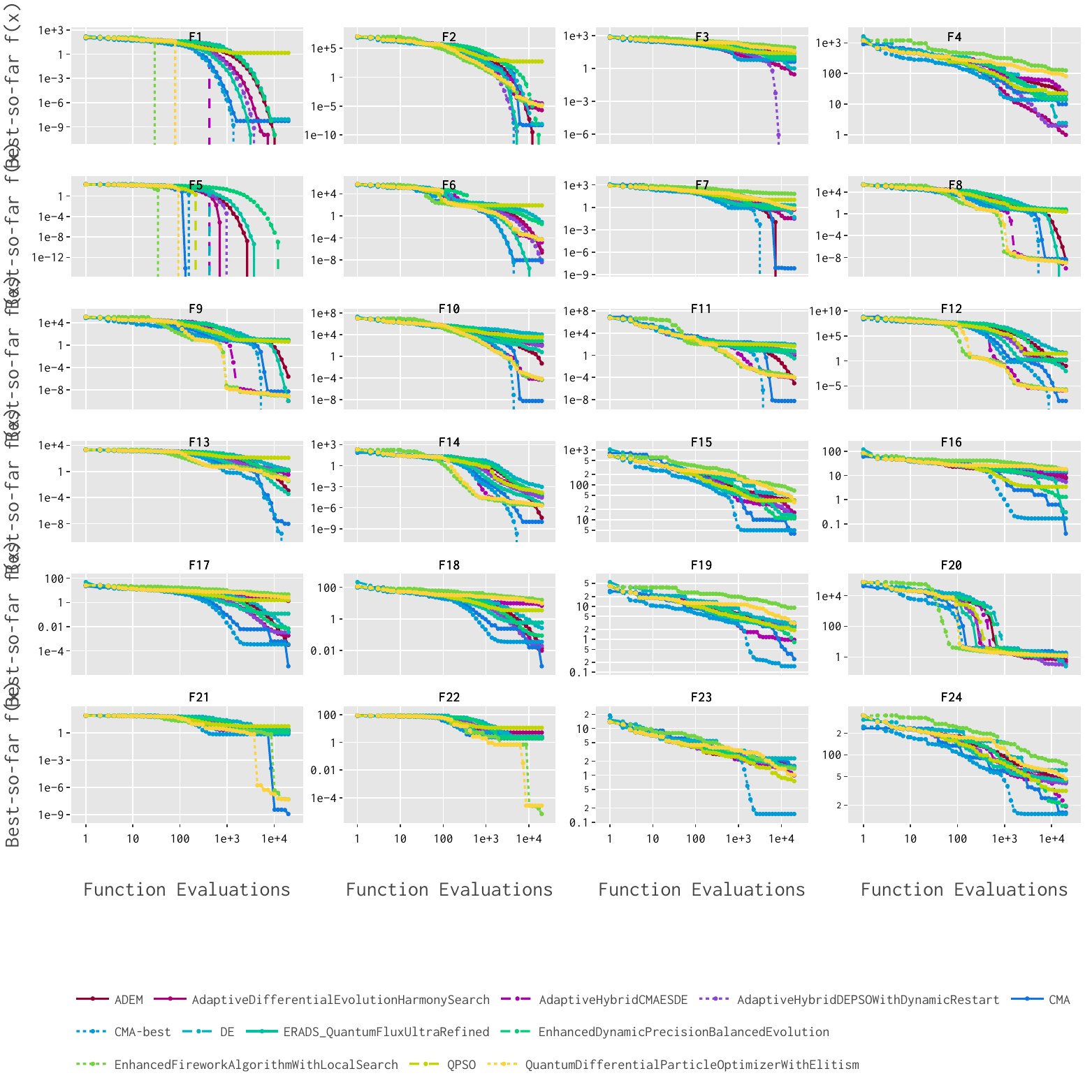}
    \caption{Median best-so-far function value ($y$-axis) over the $2\,000 \cdot d$ function evaluations ($x$-axis) per BBOB function for the best algorithm per configuration and the baselines; CMA-ES, DE and BIPOP-CMA-ES (for $d=10$).}\label{fig:FCEd10} 
\end{figure*}

\begin{figure*}[!ht]
    \centering
    \includegraphics[width=\textwidth, trim=8mm 0mm 0mm 0mm,clip]{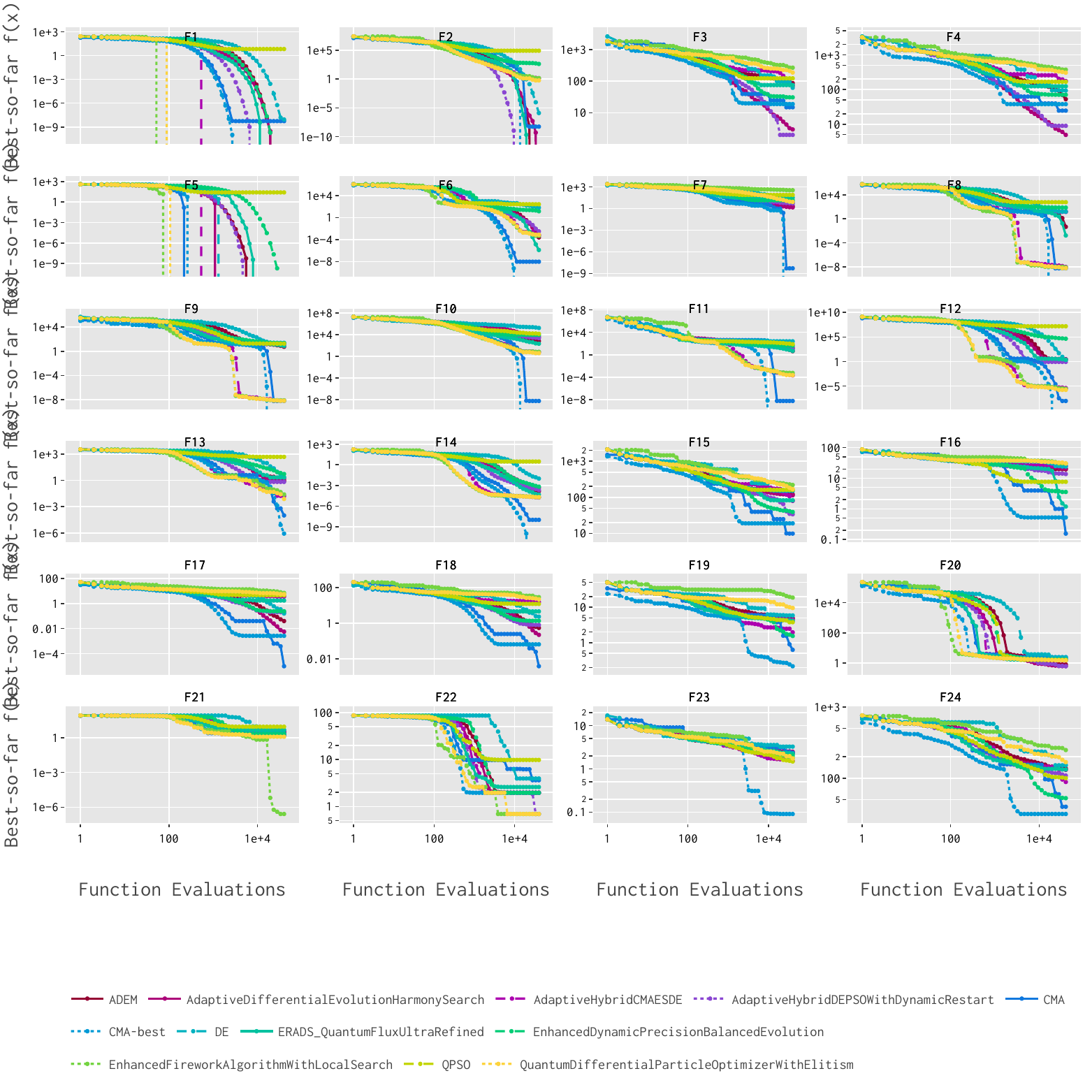}
    \caption{Median best-so-far function value ($y$-axis) over the $2\,000 \cdot d$ function evaluations ($x$-axis) per BBOB function for the best algorithm per configuration and the baselines; CMA-ES, DE and BIPOP-CMA-ES (for $d=20$).}\label{fig:FCEd20} 
\end{figure*}

\FloatBarrier
\section{Code Difference Examples}

In this section we provide a few examples of how the LLM ``mutated" a parent solution into an offspring. We give one example in Figure \ref{fig:codesnip1}, where we can observe both hyper-parameter tuning and a new control flow (new mutation strategy). 
In Figure \ref{fig:codesnip2} we can observe a mutation that only involved tuning the hyper-parameters, and in Figure \ref{fig:codesnip3} we can observe a mutation in which the crossover operator of the solution was altered.

\label{secA3}
\begin{sidewaysfigure*}[!ht]
    \centering
    \includegraphics[width=\textwidth, trim=8mm 0mm 0mm 0mm,clip]{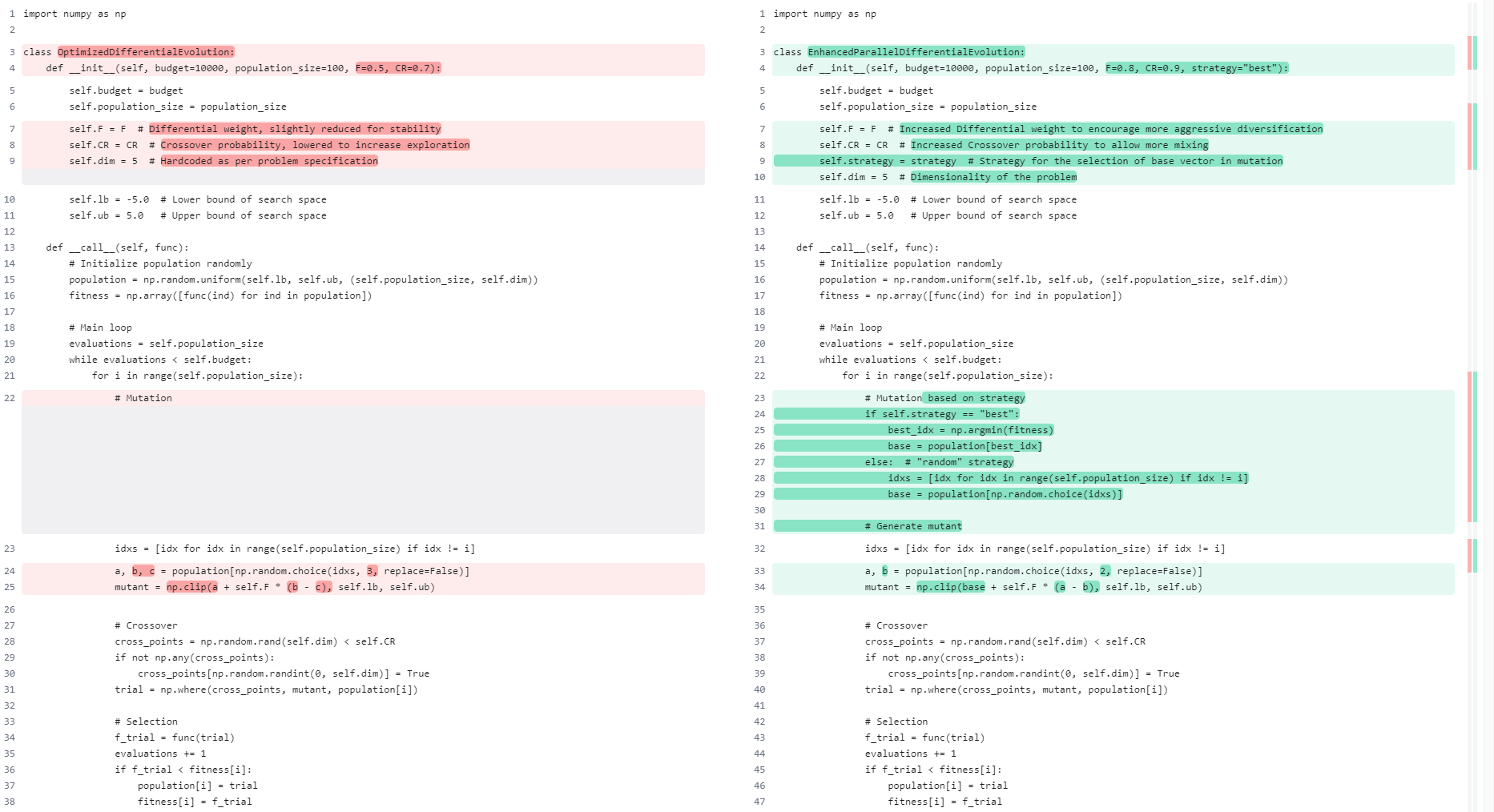}
    
    \caption{A code diff example snippet to show how the LLM mutates the parent into an offspring. This diff shows a variation of hyper-parameters (tuning) and the arguments in comments the LLM uses to change them, including a new mutation strategy. This change occurred at the beginning of the evolution process (early iteration).}\label{fig:codesnip1} 
\end{sidewaysfigure*}

\begin{sidewaysfigure*}[!ht]
    \centering
    \includegraphics[width=\textwidth, trim=8mm 0mm 0mm 0mm,clip]{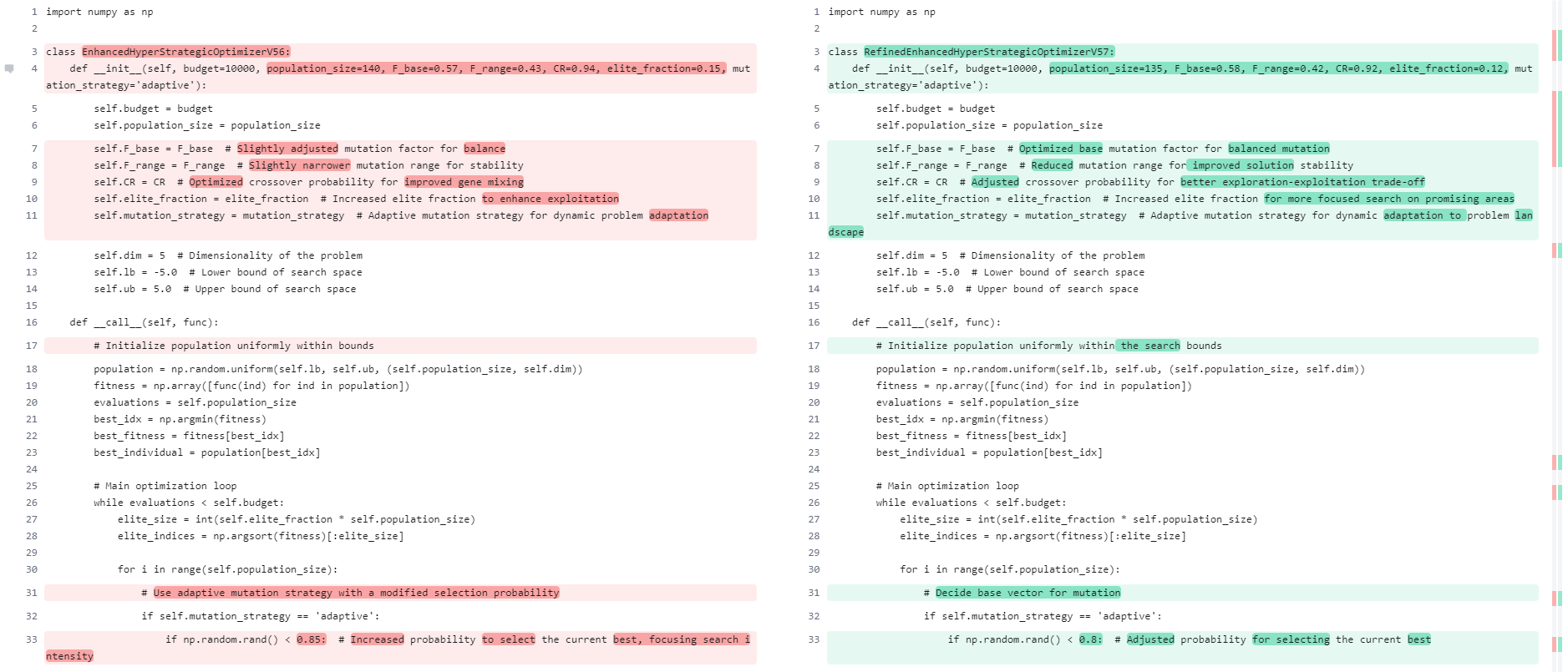}
    
    \caption{A code diff example snippet to show how the LLM mutates the parent into an offspring. This diff shows how the LLM fine-tunes some of the hyper-parameters, this change occurred at the end of the evolution process, showing a small mutation (the remaining 30 lines of code, line 34 and further, were unchanged).}\label{fig:codesnip2} 
\end{sidewaysfigure*}

\begin{sidewaysfigure*}[!ht]
    \centering
    \includegraphics[width=\textwidth, trim=8mm 0mm 0mm 0mm,clip]{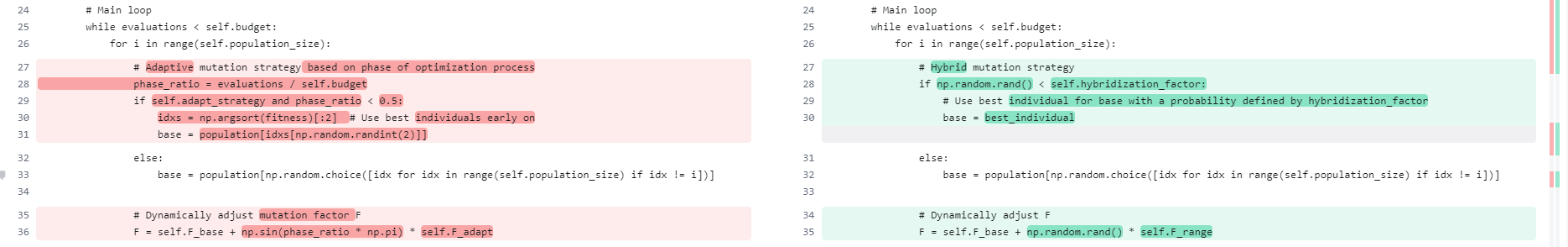}
    
    \caption{A code diff example snippet to show how the LLM mutates the parent into an offspring. This diff varies the crossover operator of the algorithm to a new approach.}\label{fig:codesnip3} 
\end{sidewaysfigure*}


\FloatBarrier

\section{Hyper-parameter Optimization of ERADS\_QuantumFluxUltraRefined}

In this supplemental section we dive deeper into the hyper-parameters of the best discovered algorithm; "ERADS\_Quantum\-Flux\-UltraRefined".
We do this by applying a hyper-parameter optimization tool, in this case the commonly used SMAC3 \cite{smac3} to tune the hyper-parameters of "ERADS\_QuantumFluxUltraRefined" on BBOB problems in $5$, $10$ and $20$ dimensions. We apply SMAC3 with a budget of $5\,000$ evaluations. The resulting hyper-parameter settings are then evaluated using all $24$ BBOB functions with $5$ instances and $5$ repetitions. The best hyper-parameters found by SMAC3 are provided in Table \ref{tab:params}.

\begin{table*}[!ht]
\centering
\caption{Comparison of default and optimized hyper-parameters of ERADS\_QuantumFluxUltraRefined for 5d, 10d, and 20d. \label{tab:params}}
\begin{tabular}{l|c|ccc}
\toprule
\textbf{Parameter}       & \textbf{Default} & \textbf{Optimized 5d} & \textbf{Optimized 10d} & \textbf{Optimized 20d} \\ 
\midrule
\textbf{Population size} & 50               & 48                    & 77                    & 75                    \\
\textbf{CR}              & 0.95             & 0.9893                & 0.8511                & 0.8556                \\
\textbf{$F_{init}$}      & 0.55             & 0.7469                & 0.5500                & 0.5605                \\
\textbf{$F_{final}$}       & 0.85             & 0.1636                & 0.7587                & 0.7317                \\
\textbf{Memory factor}   & 0.3              & 0.1043                & 0.7501                & 0.7534                \\
\bottomrule
\end{tabular}
\end{table*}

The resulting optimized algorithms are then compared against the algorithm with default (LLM generated) settings. In Figure \ref{fig:hpo5d}, the EAF curves for the default and optimized versions of ERADS\_QuantumUltraRefined can be observed. Note that only in the $20d$ case, the found optimal hyper-parameters by SMAC3 are actually outperforming the default parameters generated by the LLM. In the $5d$ and $10d$  cases the algorithms are not significantly different performing. 
From this and Table \ref{tab:params}, we can conclude that the default hyper-parameters are chosen very well, and that the $F_{init}$ and $F_{end}$ are not very sensitive parameters. It seems that for higher dimensions the memory factor should be set relatively high and the population size should also grow when the number of dimensions increases, which is common in Differential Evolution.

\begin{figure*}[!ht]
    \centering
    \includegraphics[width=0.48\textwidth, trim=0mm 20mm 0mm 0mm,clip]{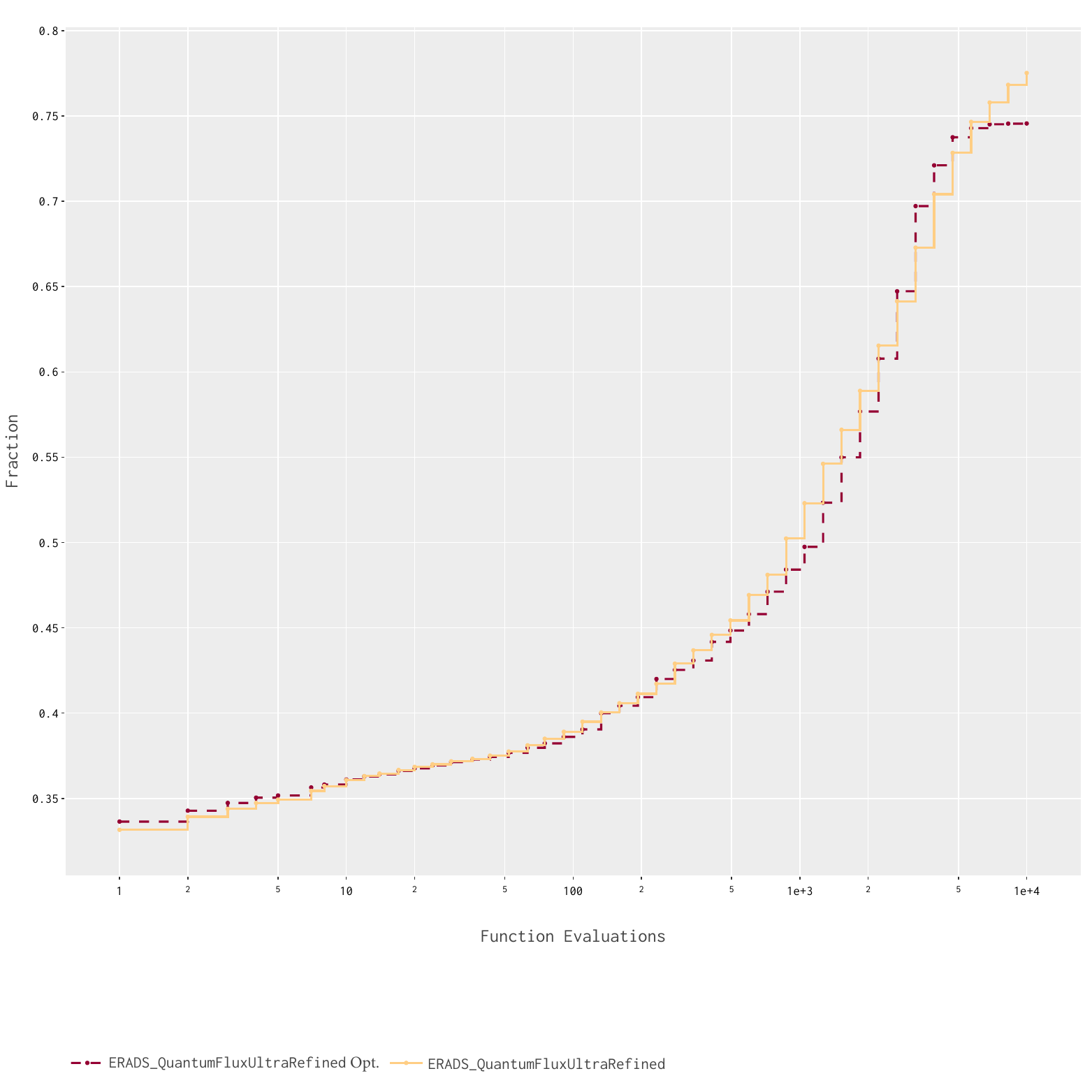}
    
    \includegraphics[width=0.48\textwidth, trim=0mm 20mm 0mm 0mm,clip]{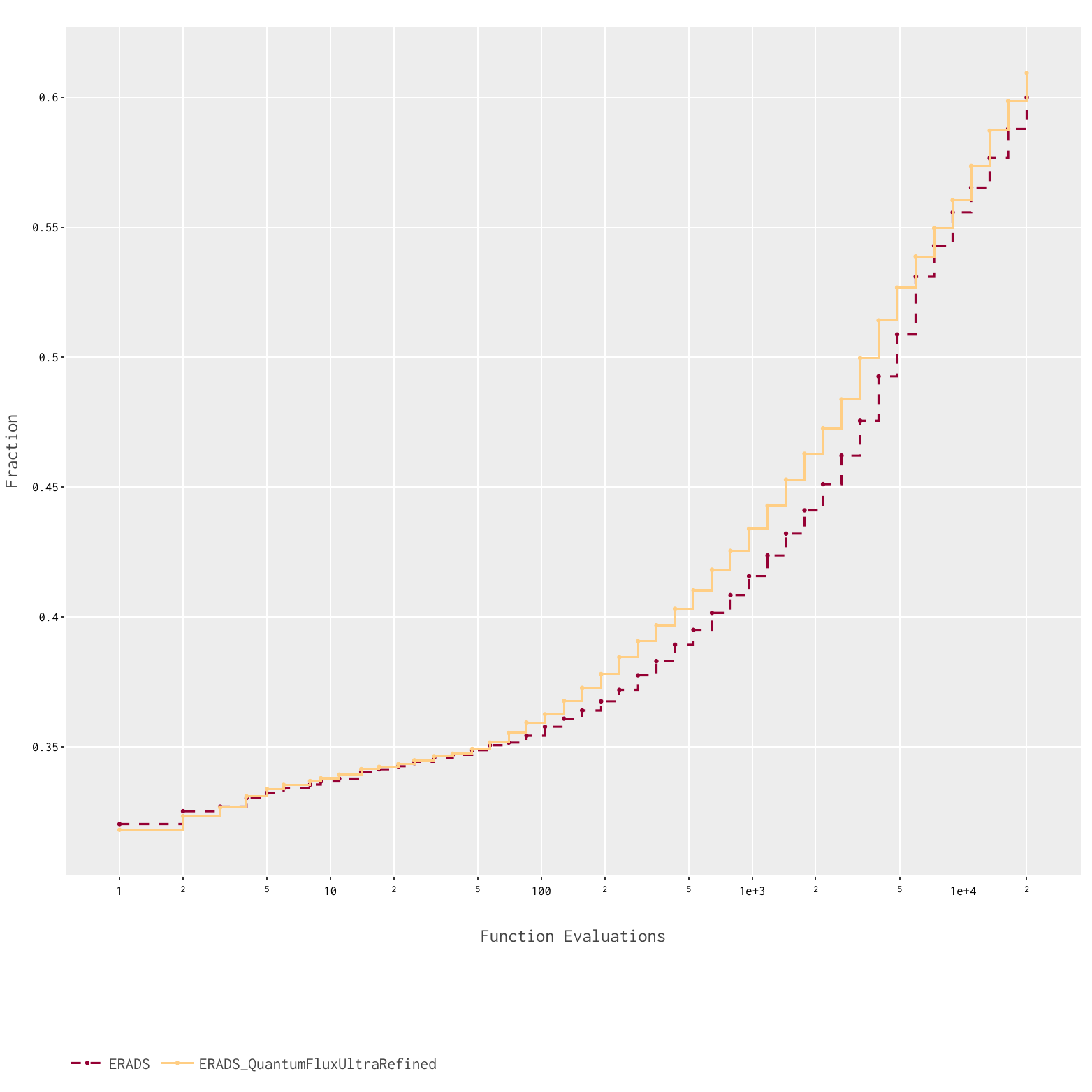}
    \includegraphics[width=0.48\textwidth, trim=0mm 20mm 0mm 0mm,clip]{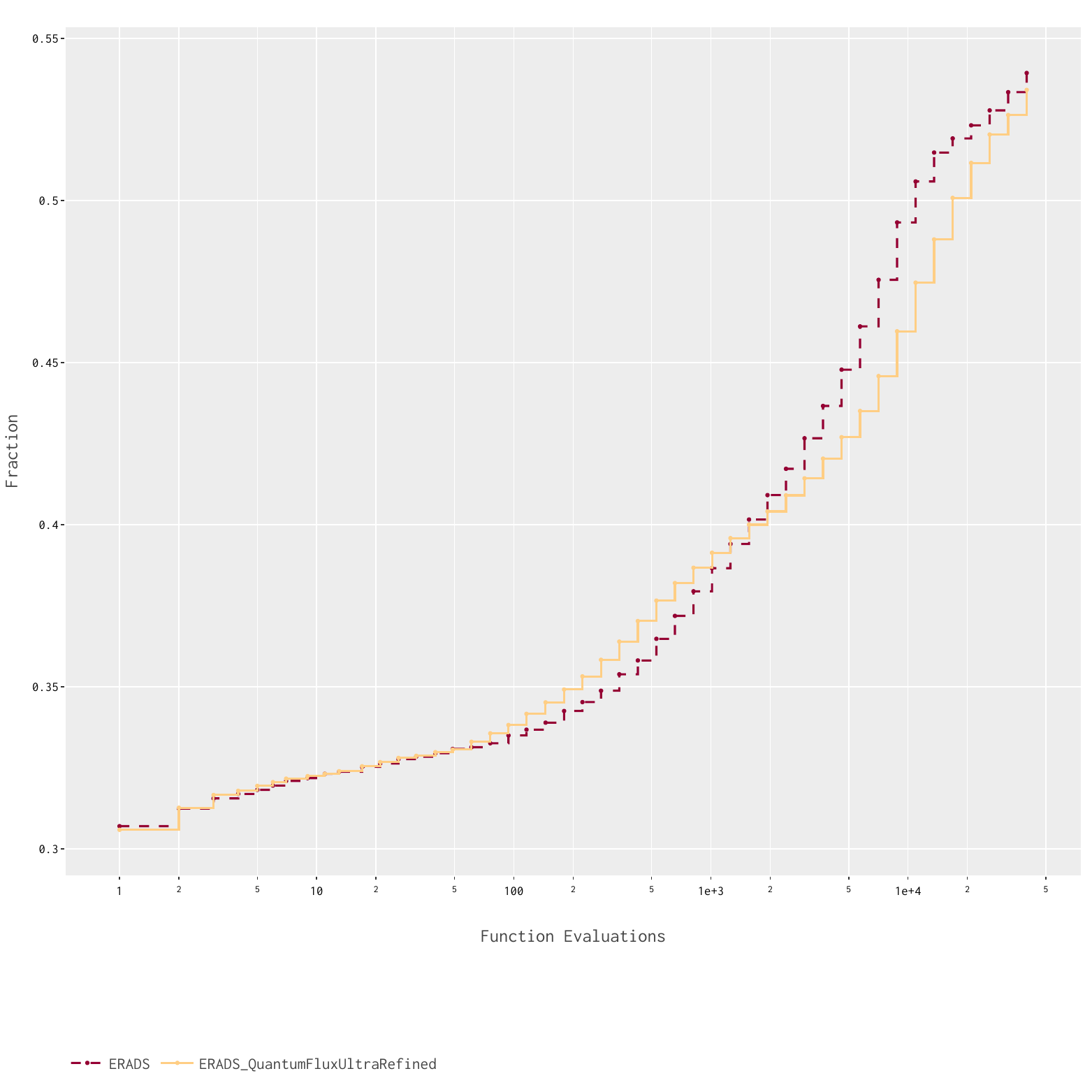}
    \includegraphics[width=\textwidth, trim=0mm 0mm 0mm 250mm,clip]{hpo/5d.pdf}
    \caption{The empirical attainment function (EAF) estimates the percentage of runs that attain an arbitrary target value not later than a given runtime. EAF for the
ERADS\_QuantumUltraRefined algorithm with default (yellow) and optimized (red) hyper-parameter settings over all 24 BBOB functions in 5d (top plot), 10d (left bottom
plot) and 20d (right bottom plot), respectively.}\label{fig:hpo5d} 
\end{figure*}
}

 





\end{document}